\documentclass[10pt,twocolumn,letterpaper]{article}
\usepackage{array}
\usepackage[pagenumbers]{cvpr}
\usepackage{microtype}
\usepackage{listings}
\usepackage{placeins}
\usepackage{algorithm}
\usepackage{algorithmic}

\definecolor{cvprblue}{rgb}{0.21,0.49,0.74}
\usepackage[pagebackref,breaklinks,colorlinks,allcolors=cvprblue]{hyperref}
\hypersetup{pdfauthor={},pdftitle={SceneAgent: A Multi-Agent Framework for Portable Game Scene Generation}}
\def\confName{CVPR}
\def\confYear{2027}
\title{SceneAgent: A Multi-Agent Framework for Portable Game Scene Generation}
\author{Xingyi Li$^{1}$\footnotemark[1]~\hspace{0.1in} 
        Jiahao Yang$^{2}$\footnotemark[1]~\hspace{0.1in} 
        Yiqin Zhu$^{1}$\hspace{0.1in}
        Jingwen Ye$^{1}$\footnotemark[2]~\hspace{0.1in} 
        Yang Sun$^{1}$\hspace{0.1in}
        Weidong Zhang$^{1}$\footnotemark[2]\\
$^1$Tencent\hspace{0.2in}$^2$Fudan University\\
{\tt\small \{xingyili,jingwenye,wadewdzhang\}@tencent.com}\\
{\small{\url{https://xingyi-li.github.io/sceneagent/}}}
\vspace{-2mm}
}
\begin{document}
\maketitle
\renewcommand{\thefootnote}{\fnsymbol{footnote}} 
\footnotetext[1]{Equal contribution.}
\footnotetext[2]{Corresponding authors.}
\begin{abstract}
Moving a scene generator to a new game engine requires more than exporting
its final output: planning, construction tools, and stored experience may
also depend on the original backend. We present
\textnormal{\textsc{SceneAgent}}, a multi-agent framework for reusing the
construction process and its experience across backends. A canonical
SceneContext maintains the evolving world, while backend adapters isolate
coordinate conventions, asset bindings, and export. Structured handoffs let
fresh Builder sessions query this persistent state without inheriting the
full construction transcript. Across generations, human-approved traces are
distilled into asset-ID-free Layout skills and semantic Content skills that
can be retrieved and recombined. We introduce GameSceneBench, comprising 69
indoor and outdoor prompts. Under a common asset and evaluation protocol,
\textnormal{\textsc{SceneAgent}} obtains human tie-adjusted preference scores
of $78.2$--$89.4\%$ against five baselines. Ablations show that reusable skills
and the multi-agent protocol primarily improve prompt alignment and scene
completeness, with additional runtime. Deployment in a second engine
preserves the shared construction protocol and skill library while replacing
backend integration components. Together, these results support separating
persistent scene state, task-local reasoning, and reusable experience for
portable scene construction.
\end{abstract}

\section{Introduction}
\label{sec:intro}

Constructing a 3D game environment requires coordinated design decisions:
buildings must follow roads, landmarks must face useful viewpoints, and
vegetation must respect occupied space. These decisions should remain useful
when the engine or asset library changes. Existing LLM-based
systems can construct complex scenes~\cite{songtang2025unrealllm,duan2025latticeworld},
but construction decisions are often expressed through backend-specific
coordinates, asset identifiers, APIs, and execution traces. An interoperable
output file alone does not make these decisions reusable.

We study \emph{construction portability}: retaining the planning protocol,
scene-state interface, and reusable design knowledge while localizing
backend-specific implementation. This requires more than a common export
schema. During generation, agents need to coordinate against the same
evolving world without inheriting every preceding tool call. Across
generations, experience must retain useful spatial and semantic patterns
without carrying over incidental asset bindings. Both world state within a run and design knowledge across runs must remain
accessible through the shared interface.

We introduce \textsc{SceneAgent}, organized around three complementary
mechanisms.

\textbf{Canonical construction state.}
A shared \emph{SceneContext} records actor transforms, bounding boxes,
occupied regions, and subtask provenance in canonical coordinates. Builders
query and update this state through construction tools. Adapters handle
coordinate conversion, asset bindings, and export. Backend-specific
capability wrappers expose native operations through the construction
interface. This boundary preserves the shared protocol while making the
required backend integration explicit.

\textbf{Structured handoffs.}
An Orchestrator launches a fresh Builder session for each planned subtask. The Builder receives its objective, shared conventions, relevant skills, and compact completion reports and then queries SceneContext for additional world information. Thus, later subtasks can respect earlier construction without replaying its full transcript. Geometry checks and bounded
refinement provide control points before the next handoff.

\textbf{Factored experience.}
Human-approved traces are distilled into asset-ID-free \emph{Layout} skills
for spatial organization and semantic \emph{Content} skills for object roles,
style, and substitution. Thin recipes compose these factors with
prompt-specific requirements. Backend databases supply concrete assets.
The same route layout can therefore guide different scene themes, and the
same theme can be realized through different layouts. Rejected traces are
retained separately as failure evidence.

We evaluate \textsc{SceneAgent} on GameSceneBench, a benchmark of 69 indoor
and outdoor prompts. Under a shared asset and rendering protocol, a blinded
human study yields tie-adjusted preferences of $78.2$--$89.4\%$ over five
baselines. VLM comparisons favor the method in alignment, spatial
plausibility, and completeness. Skill-library and single-agent ablations
show the clearest gains in alignment and completeness, while quantifying
the runtime cost. A second-engine deployment tests reuse of the shared
protocol and skills, and cross-LLM studies examine model dependence.

Our contributions are:
\begin{itemize}
\item A construction-portability design that couples canonical scene state
with a shared tool interface and an explicit backend integration boundary.
\item A Planner--Builder protocol that coordinates fresh task-local sessions
through persistent scene state and structured completion reports.
\item A factored Layout--Content experience library for compositional reuse,
with evaluation on GameSceneBench through baseline comparisons, ablations,
and cross-engine and cross-model studies.
\end{itemize}

\section{Related Work}
\paragraph{Agentic Scene Generation and Portable Representations.}
Language models have expanded text-to-scene generation from layout prediction~\cite{feng2023layoutgpt} to assembling and refining complete 3D scenes~\cite{berdoz2026text}. Holodeck~\cite{yang2024holodeck} populates scenes with retrieved assets, while Holodeck 2.0~\cite{bian2025holodeck} adds generated assets and interactive editing. Other agentic systems combine specialized roles, tool use, and visual feedback~\cite{ling2025scenethesis,yang2026sceneweaver,luo2026sceneassistant}, while SAGE~\cite{xia2026sage} further introduces visual and physics critics for simulator-validated generation. The Procedural Scene Programs approach~\cite{gumin2025procedural} instead represents scene layouts as imperative programs, placing objects relative to previously constructed elements and improving layout validity through LLM-free program search. Many existing systems realize these capabilities through specific authoring environments: SceneCraft~\cite{hu2024scenecraft} and CityX~\cite{zhang2024cityx} generate Blender programs or procedural structures, whereas UnrealLLM~\cite{songtang2025unrealllm}, AutoUE~\cite{yin2026autoue}, and LatticeWorld~\cite{duan2025latticeworld} rely on Unreal Engine facilities and tooling.

A complementary line of work develops structured scene representations that remain editable and renderable across graphics systems. The Scene Language~\cite{zhang2025scene} represents structure, semantics, and visual identity through programs, words, and embeddings, supporting traditional, neural, and hybrid renderers. GenUSD~\cite{lin2024genusd} hierarchically plans scene layouts, generates 3D assets on demand, assembles them in Omniverse, and exports the resulting scenes in USD, thereby providing an interoperable scene output. \textsc{SceneAgent} focuses on portability of the construction process:
agents query and update canonical world state through a shared interface,
while adapters and capability wrappers handle backend realization. This
complements interoperable output representations by also retaining the
planning protocol and semantic experience across backend deployments.

\paragraph{Long-Horizon State and Reusable Experience.}
Long-horizon agents increasingly externalize task state instead of relying on the full interaction history. Magentic-One~\cite{fourney2024magentic} coordinates specialized agents through structured task and progress ledgers, while InfiAgent~\cite{yu2026infiagent} reconstructs bounded reasoning contexts from persistent workspace state and a fixed window of recent actions. Another line of work distills prior trajectories into reusable knowledge: Voyager~\cite{wang2023voyager} stores successful behaviors as executable skills, ExpeL~\cite{zhao2024expel} extracts transferable insights from trajectories, and Agent Workflow Memory~\cite{wang2024agent} induces reusable workflows. In 3D scene synthesis, SceneMotifCoder~\cite{tam2025scenemotifcoder} similarly distills example arrangements into reusable visual meta-programs.

\textsc{SceneAgent} brings these two forms of persistence into one
construction protocol. SceneContext carries the evolving world between
Builder sessions, while approved traces supply spatial and semantic guidance
for later scenes. Layout and Content are stored and retrieved separately,
with concrete asset mappings left to backend databases. The distinction is
between reusable scene-design guidance and the backend-bound details of an
individual construction trace.

\section{System Overview}
\begin{figure*}[t]
    \centering
    \includegraphics[width=\textwidth]{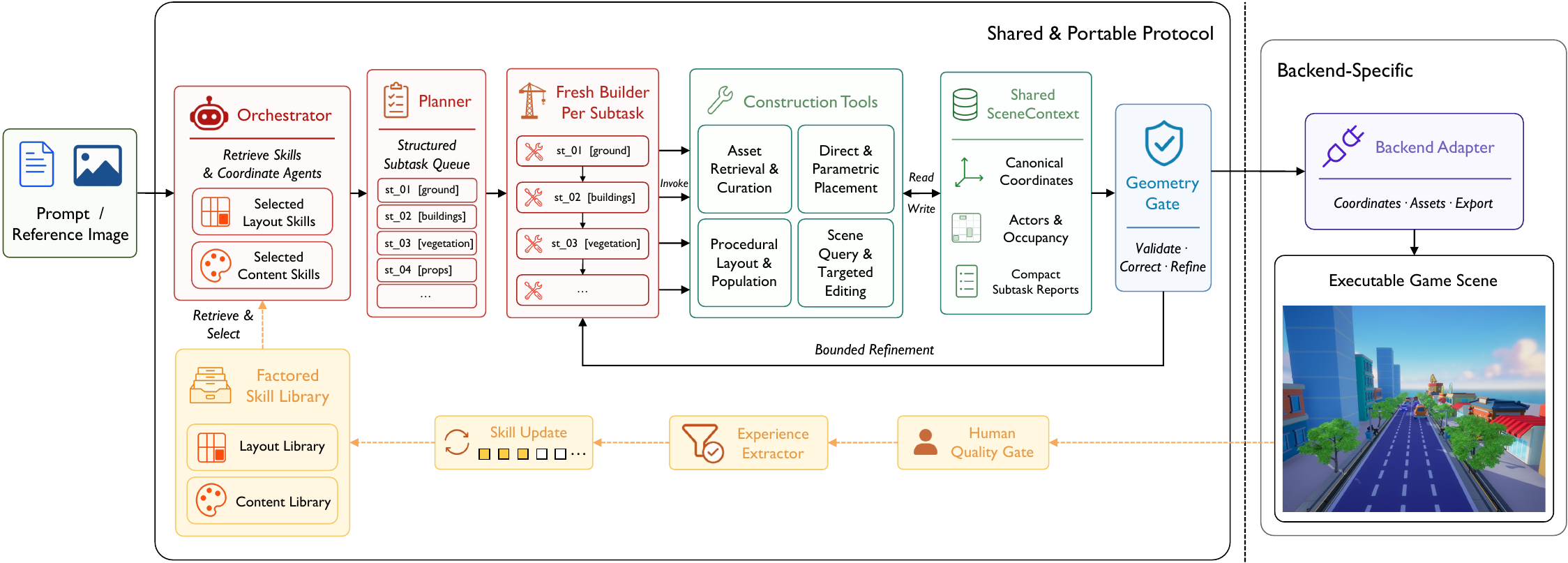}
    \caption{\textsc{SceneAgent} separates within-run construction from
    cross-run experience reuse. Fresh Builders coordinate through canonical
    scene state. Backend adapters realize that state in the target engine.
    The optional human-gated update path retains approved traces as
    composable Layout and Content skills.}
    \label{fig:overview}
\end{figure*}

Figure~\ref{fig:overview} shows the generation and experience-update paths.
Given a prompt or reference image, the Orchestrator retrieves Layout and
Content skills and supplies them as read-only guidance to the Planner and
relevant Builders. The Planner produces an ordered queue of semantic
subtasks. For each subtask, a fresh Builder reads its task record and compact
completion reports, queries SceneContext, and modifies the world through
construction tools. Geometry checks diagnose the updated state and support
bounded refinement. After construction, the backend adapter exports the
canonical state to the target engine.

Two forms of information persist beyond individual Builder sessions.
SceneContext stores the current world within a generation, while the skill
library stores reusable guidance across generations. Human review determines
which completed traces enter the experience-update path. Automatic checks
provide supporting diagnostics. Planning, handoffs, canonical state, and
semantic skills are shared across backends. Asset configuration, export
adapters, and native-capability wrappers supply backend-specific realization.
The Orchestrator coordinates these stages without placing actors directly.

\section{Method}
\label{sec:method}
\textsc{SceneAgent} separates three responsibilities: storing the constructed
world, reasoning about the current subtask, and retaining experience for
future scenes.

SceneContext records what has been built, handoffs summarize task outcomes
and unresolved issues, and skills provide reusable design guidance. A fresh
Builder combines these sources without inheriting the full construction
transcript. This division lets task-local reasoning use persistent world
state and cross-scene experience, while adapters and asset databases handle
backend-specific realization.

\subsection{Canonical Scene State and Backend Boundary}
The Planner expresses scene intent through semantic categories, roles, spatial relations, constraints, and gameplay affordances. Builders realize this intent in a shared SceneContext, rather than by invoking an engine directly. SceneContext stores each concrete actor in a builder coordinate system ($X$: left--right, $Y$: up, $Z$: forward--back), together with its asset reference, transform, bounding box, source tag, and subtask identifier. It also maintains scene bounds and an occupancy index for spatial queries. SceneContext serves as both an exportable world representation and the live
coordination state: later Builders query what earlier subtasks constructed,
including occupied space and provenance, without replaying their actions.

The exported payload is actor-centric: each actor carries an asset reference, a transform, and optional appearance or collision metadata. A backend adapter converts SceneContext into the target backend's coordinate, unit, scale, and rotation conventions, resolves backend-specific asset bindings when required, and serializes the result. For example, one adapter swaps the builder $Y$ and $Z$ axes, converts game units to meters, and emits quaternion rotations. The same construction state can therefore be exported through
another adapter while preserving how Builders place actors
and query the scene.

Table~\ref{tab:portability-boundary} defines the portability boundary.
Retargeting requires a compatible asset catalog, an adapter, and wrappers for
backend-specific capabilities. The canonical coordinates, SceneContext
schema, core tool interface, Planner--Builder protocol, and semantic skill
libraries remain shared. Portability therefore concerns reuse of the
construction process with explicit backend integration.

\paragraph{From semantic design to backend assets.}
Retargeting preserves semantic roles and spatial rules while grounding them
in the active backend. For a street scene, a Layout skill can specify a
central route with buildings along its sides, while Content skills specify
building families and supporting props. Builders retrieve candidates from
the local catalog, inspect their metadata and dimensions, and place selected
assets relative to SceneContext. Content skills can guide substitution when
a requested object is unavailable. Placement and scene queries use the shared
coordinate conventions. The adapter then converts transforms and resolves
backend-specific bindings for export. This process reuses the construction
strategy while allowing concrete assets and scene geometry to reflect the
target catalog.

\begin{table}[t]
\caption{Shared and backend-local components in \textsc{SceneAgent}.}
\label{tab:portability-boundary}
\centering
\footnotesize
\begin{tabular}{
  @{}
  >{\raggedright\arraybackslash}p{0.34\columnwidth}
  >{\raggedright\arraybackslash}p{0.60\columnwidth}
  @{}
}
\toprule
Component & Responsibility \\
\midrule
\multicolumn{2}{@{}l}{\textit{Shared across backends}} \\
Agent protocol & Task decomposition and handoffs. \\
SceneContext & Canonical state and spatial queries. \\
Reusable skills & Layout and Content guidance. \\
\midrule
\multicolumn{2}{@{}l}{\textit{Backend-specific integration}} \\
Asset configuration & Asset catalogs and semantic mappings. \\
Backend adapter & Coordinate conversion, asset binding, and export. \\
Tool wrappers & Engine operations under shared conventions. \\
\bottomrule
\end{tabular}
\end{table}

\subsection{Planning and Tool-Mediated Construction}
The Planner transforms the prompt and an optional reference image into an ordered queue of subtasks, such as ground, building, vegetation, prop, and refinement tasks. Each record specifies a semantic objective, target region, candidate content, construction guide, and shared conventions including the main axis, orientation rules, scale unit, and ground height.

For each subtask, the Orchestrator launches a fresh Builder session. Its initial context includes the current task record, construction guide, shared conventions, selected Layout and Content skills, and compact reports from completed subtasks, while excluding raw placement and tool-call histories. When additional information is required, the Builder queries the live SceneContext as needed instead of reconstructing the world from a transcript.

The shared construction tools cover asset retrieval and inspection, direct
and procedural placement, scene queries, and targeted editing. Geometric
operations use builder coordinates and read from or write to SceneContext.
For example, Builders can inspect candidate assets, populate a region, query
per-subtask footprints, or remove actors from an occupied area. Additional
backend-local wrappers encapsulate native APIs and normalize inputs and
outputs to shared conventions where possible. The supplementary material
lists the construction and agent-control interfaces.

Scene-changing operations trigger geometry checks for collisions, unsupported or out-of-bounds actors, extreme transforms, invalid assets, and density anomalies. High-confidence violations can receive conservative corrections, such as removal, clamping, or re-seating. Unresolved cases trigger a bounded number of refinement passes or return control to the user. At the end of a subtask, the Builder returns a compact report containing its completion status, summary, residual issues, and actor-count change. SceneContext, rather than the Builder conversation, carries the constructed world into the next subtask.

For a night market, the street Builder first establishes the road. The
market-stall Builder receives its completion report and queries SceneContext
for road geometry and occupied regions before placing vendors along the
sides. A later prop Builder can inspect the resulting footprints before
adding furniture. Each handoff preserves task outcomes and access to the
world while discarding the preceding Builder's raw tool-call history.

\subsection{Factored Experience Representation}
A stored scene trace can entangle spatial organization, semantic content,
and concrete asset choices. To make these decisions independently reusable,
\textsc{SceneAgent} extracts three complementary objects:
\begin{itemize}
    \item \textbf{Layout skill}: reusable spatial organization and geometric constraints, such as grouping, support, symmetry, focal placement, path structure, density gradients, and terrain constraints. It contains no concrete asset identifiers.
    \item \textbf{Content skill}: semantic roles, stylistic attributes, and substitution rules, such as object families, theme tags, and material styles. It contains no backend-specific bindings.
    \item \textbf{Thin recipe}: a scene-specific composition record that references Layout and Content skills and retains only residual details unique to the current prompt.
\end{itemize}

Factoring allows the same spatial organization to be paired with different content families and the same content family to be realized through different layouts, without storing a separate fused memory for every combination. A campsite layout, for example, may be paired with medieval, science-fiction, or cartoon content, while a village theme may use grid, radial, or terrain-following organization. 
This factorization need not perfectly decompose every scene,
but keeps concrete asset IDs and backend directives separate
from transferable experience.

\paragraph{Example of skill composition.}
Consider a night-market scene organized around a pedestrian street.
A linear Layout skill specifies a continuous route between an entrance
and a terminal focus, places major regions along its sides, and preserves
an unobstructed circulation corridor. A night-market Content skill supplies
semantic roles such as vendors, lighting, furniture, and supporting props,
together with stylistic guidance and substitution rules. For example, a
missing food stall can be replaced by a canopy-and-counter assembly.
A thin recipe combines these skills and adds scene-specific requirements,
such as staggered vendor rows and small dining pockets. Concrete assets
are subsequently resolved through the backend-specific asset database.
Thus, reusable spatial rules, semantic content, and prompt-specific
requirements remain separately represented.

\subsection{Quality-Gated Experience Update}
Experience is retained through a human-led quality gate, with automatic checks serving as supporting diagnostic evidence. After generation, \textsc{SceneAgent} packages the prompt, construction record, final SceneContext, multi-view renderings, automatic validation results, and checkpoint feedback into an experience candidate. 
The automatic checks expose collisions, unsupported actors,
invalid transforms, and density anomalies, but do not alone
determine whether to retain the experience.

A human reviewer makes the final decision by considering prompt adherence, spatial organization, visual coherence, richness, and potential reusability. The reviewer may approve or reject the trace and optionally provide a natural-language critique or correction. Only approved traces are passed to the experience extractor, which assigns transferable spatial rules to Layout skills, semantic and stylistic knowledge to Content skills, and prompt-specific residuals to a thin recipe. Candidate skills then undergo schema validation and deduplication before insertion into the library for future reuse. 
Rejected traces and reviewer feedback are retained separately
as failure evidence.

At generation time, the Orchestrator retrieves a compact set of Layout and Content skills from their indexes and injects them as structured, read-only sections of the system prompts for the Planner and relevant Builder sessions. 
Separate indexes enable independent retrieval of spatial and
semantic experience, while suppressing duplicate loads.

\paragraph{Visual experience acquisition.}
Reference images or screenshots provide an additional source of reusable
experience. A vision-language model interprets their semantic content,
spatial relations, and construction cues, which then condition the standard
Planner--Builder pipeline. The resulting scene and construction trace undergo
the same human-led quality gate as text-prompted generations. Only approved
traces are factorized into Layout and Content skills and inserted into the
experience library after schema validation and deduplication. This route can
incrementally expand the library from visual examples without changing the
skill representation, retrieval mechanism, or update protocol.

\begin{figure*}[t]
    \centering
    \includegraphics[width=\textwidth]{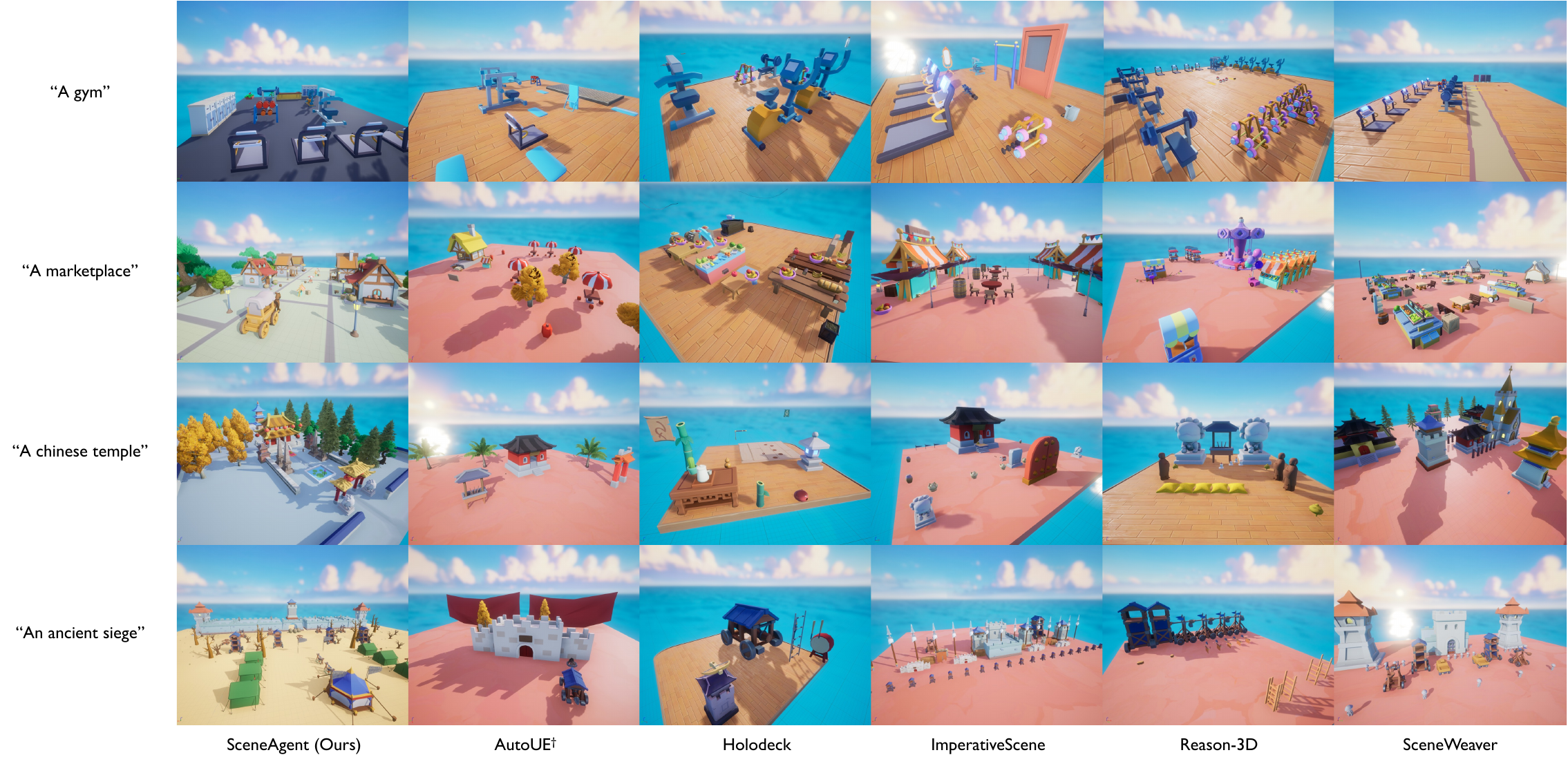}
    \caption{Qualitative comparison on representative GameSceneBench prompts.
    Rows share an input prompt, and columns show different methods under the
    common Engine~A asset and rendering protocol.}
    \label{fig:qualitative-comparison}
\end{figure*}

\section{GameSceneBench and Evaluation Metrics}
\label{sec:evaluation}
\paragraph{GameSceneBench.}

We introduce GameSceneBench, a text-driven benchmark containing 69 prompts:
56 outdoor and 13 indoor scenes spanning diverse natural, built, historical,
stylized, and functional environments. Prompts range from open-ended requests
to detailed descriptions with required objects and spatial relations. Every
method uses the same asset catalog, rendering pipeline, and camera protocol
and is run independently three times per prompt. Each scene is rendered from
four perspective views and one top-down view. Because multiple realizations
may satisfy a prompt, we evaluate alignment, efficiency, spatial plausibility,
and completeness rather than reconstruction error. The prompts, category
breakdown, and rendering details are provided in the supplementary material.

\paragraph{CLIPScore and runtime.}
We compute CLIPScore~\cite{hessel2021clipscore} with CLIP
ViT-L/14~\cite{radford2021learning} for each prompt--view pair and average over
the five views.
Runtime measures wall-clock time from prompt submission to scene completion,
including planning and construction but excluding offline rendering, VLM
evaluation, and human evaluation.
Both metrics are
averaged over the three runs.

\paragraph{VLM evaluation.}
We evaluate \emph{text--scene alignment} (prompt satisfaction), \emph{spatial
plausibility} (placement, support, scale, accessibility, and overlap), and
\emph{scene completeness} (sufficient, nonredundant supporting content). For
each method pair, prompt, and run, the two sets of five views form an anonymous
$2\times5$ composite with randomized top--bottom order. The VLM selects top,
bottom, or tie separately for each dimension. We repeat the comparison with
independently randomized ordering and report tie-adjusted preference, counting
a tie as half a win. For method $A$, the preference score is
\begin{equation}
    P_A = \frac{W_A + 0.5T}{W_A + L_A + T}\times100\%,
    \label{eq:preference}
\end{equation}
where $W_A$, $L_A$, and $T$ count wins, losses, and ties.
$50\%$ indicates parity. Human preferences use the same definition. These judgments assess visible
scene quality from the rendered views. They do not directly test physical
stability or successful navigation.

\paragraph{Human evaluation.}

We recruit 65 volunteers and exclude 14 who do not complete the required
number of comparisons, leaving 51 participants and 1,860 judgments. Given the
prompt and the same anonymous, randomized $2\times5$ composite, participants
select the scene with better overall plausibility and consistency with the
input text, or indicate no preference. Method identities are hidden, and no
preference counts as half a win. Study oversight, consent, and protocol details are provided in the
supplementary material.

\section{Experiments}
\subsection{Experimental Setup}
Unless otherwise stated, all experiments use Engine~A as the target backend
and follow the GameSceneBench protocol. All methods use
DeepSeek-V4-Flash~\cite{xu2026deepseek} with a context window of up to 204,800
tokens. Pairwise VLM evaluation uses \texttt{gpt-4o-2024-11-20}.
For \textsc{SceneAgent}, the Orchestrator is limited to 30 interaction rounds, the Planner
to 50 rounds and 40 tool calls, and each Builder session to 40 rounds.
The Planner produces 5--10 outdoor or 3--6 indoor subtasks,
and the Orchestrator allows up to three automatic refinement rounds.

For fairness, all baselines query the same Engine~A retrieval service and asset
catalog while retaining their own search prompts. We otherwise preserve their
original planning, prompting, and construction procedures wherever possible.
We evaluate AutoUE~\cite{yin2026autoue} through its scene-layout stage in the
common Engine~A pipeline. Full adaptation details are
provided in the supplementary material. The experience library is frozen
before evaluation and is not updated on GameSceneBench.

\subsection{Main Comparison on GameSceneBench}
We compare \textsc{SceneAgent} with AutoUE~\cite{yin2026autoue},
Holodeck~\cite{yang2024holodeck}, ImperativeScene~\cite{gumin2025procedural}, Reason-3D~\cite{berdoz2026text}, and
SceneWeaver~\cite{yang2026sceneweaver}.
Table~\ref{tab:main-comparison} reports the results.

\begin{table}[t]
\caption{GameSceneBench comparison. Panel (a) reports mean $\pm$ standard
deviation. Panel (b) reports the tie-adjusted preference for
\textsc{SceneAgent} against each comparator, with $50\%$ indicating parity.
$^\dagger$ denotes AutoUE adapted to the common backend using its original
scene-layout stage.
Detailed results are provided in the supplementary
material.}
\label{tab:main-comparison}
\centering
\footnotesize
\textbf{(a) Absolute metrics}\\[2pt]
\setlength{\tabcolsep}{4pt}
\begin{tabular}{lcc}
\toprule
Method & CLIPScore $\uparrow$ & Runtime (s) $\downarrow$ \\
\midrule
AutoUE$^\dagger$
    & $0.188 \pm 0.030$ & $\mathbf{16.3 \pm 2.9}$ \\
Holodeck
    & $0.185 \pm 0.030$ & $60.7 \pm 19.1$ \\
ImperativeScene
    & $0.214 \pm 0.030$ & $21.4 \pm 29.8$ \\
Reason-3D
    & $0.196 \pm 0.033$ & $23.2 \pm 11.4$ \\
SceneWeaver
    & $0.193 \pm 0.030$ & $226.4 \pm 140.0$ \\
\textsc{SceneAgent}
    & $\mathbf{0.216 \pm 0.034}$ & $203.2 \pm 97.8$ \\
\bottomrule
\end{tabular}
\par\smallskip
\textbf{(b) \textsc{SceneAgent} preference vs. comparator (\%)}\\[2pt]
\setlength{\tabcolsep}{4pt}
\begin{tabular}{lcccc}
\toprule
Comparator & Align. $\uparrow$ & Plaus. $\uparrow$
    & Comp. $\uparrow$ & Human $\uparrow$ \\
\midrule
AutoUE$^\dagger$  & $94.2$ & $74.4$ & $91.3$ & $89.4$ \\
Holodeck          & $81.5$ & $74.1$ & $77.1$ & $85.2$ \\
ImperativeScene   & $77.3$ & $65.3$ & $74.9$ & $78.2$ \\
Reason-3D         & $78.3$ & $61.6$ & $82.6$ & $81.2$ \\
SceneWeaver       & $83.5$ & $64.4$ & $76.7$ & $82.7$ \\
\bottomrule
\end{tabular}

\end{table}

\begin{figure}[t]
    \centering
    \includegraphics[width=\columnwidth]{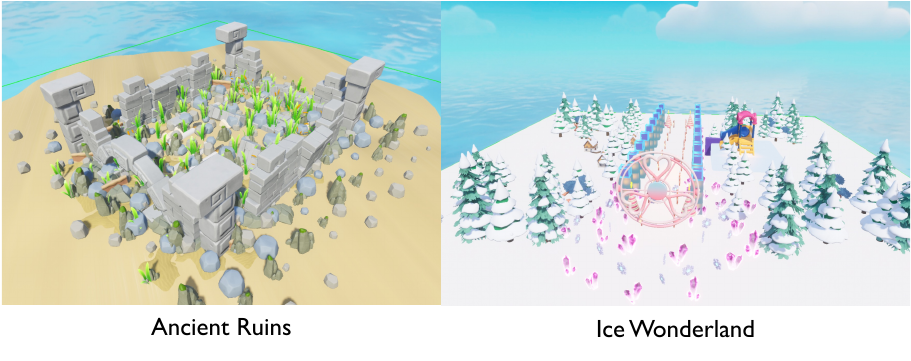}
    \caption{Representative generations in Engine~B.}
    \label{fig:cross-engine}
\end{figure}

Human evaluation consistently favors \textsc{SceneAgent}. Across 1,860 judgments,
\textsc{SceneAgent} wins $76.9\%$, ties in $12.7\%$, and loses $10.3\%$,
yielding an overall tie-adjusted preference of $83.3\%$. It is preferred to
each baseline, with scores of $78.2$--$89.4\%$. Pairwise VLM evaluation agrees
in direction: \textsc{SceneAgent} is favored over every comparator in all
three dimensions, with aggregate preferences of $72.5$--$86.7\%$.

CLIPScore is also highest at $0.216$, although the margin over ImperativeScene
($0.214$) is small. The preference results therefore provide complementary
evidence about scene organization and content beyond global text--image
similarity. These gains incur additional computation: average generation
time is $203.2$ seconds, lower than SceneWeaver but higher than the other
baselines. Figure~\ref{fig:qualitative-comparison} provides representative
outputs under the same asset and rendering protocol.

\subsection{Cross-Engine and Cross-Model Deployment}
\paragraph{Cross-engine.}
To demonstrate cross-backend portability, we deploy \textsc{SceneAgent} in Engine~B and run all 69 GameSceneBench prompts while keeping the language model, agent
protocol, SceneContext, core tools, skills, and generation budget fixed. Only
the adapter, asset catalog and database, and capability-tool wrappers change.
Figure~\ref{fig:cross-engine} shows representative Engine~B generations.
Additional outputs and export schemas appear in the supplement. This study
tests whether the shared protocol can be retargeted through the stated
backend boundary. Because the asset catalogs differ, the examples provide
qualitative evidence of retargeting rather than a matched quality comparison.

\paragraph{Cross-LLM.}
We replace the main DeepSeek-V4-Flash backbone with Gemini 3.5
Flash~\cite{gemini35flash}, Hy3~\cite{tencent2026hy3}, and
Qwen3.6-35B-A3B~\cite{qwen2026qwen36}, running each on all 69 GameSceneBench prompts with all
other settings fixed.

\begin{figure}[t]
    \centering
    \includegraphics[width=\columnwidth]{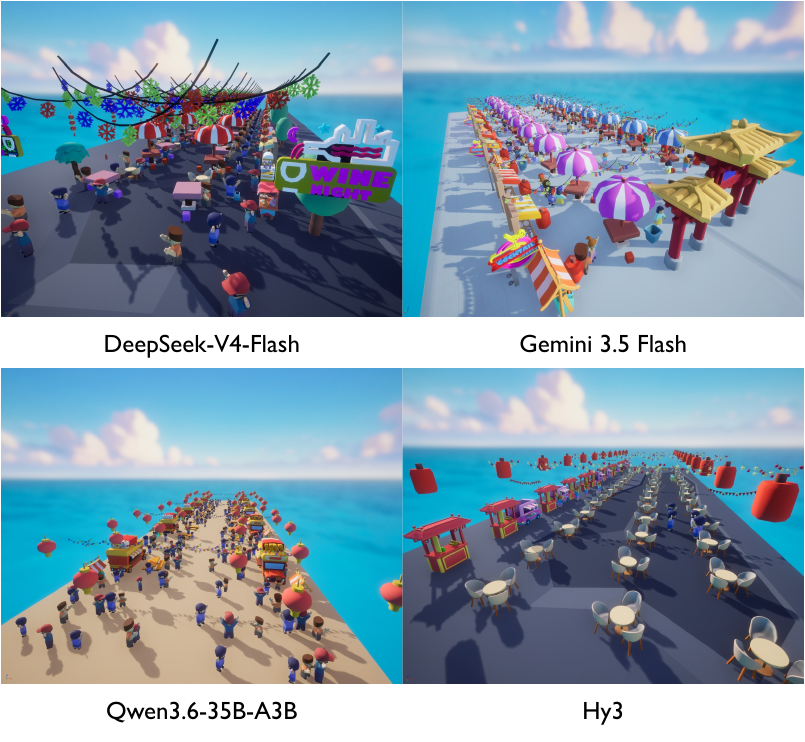}
    \caption{Cross-LLM comparison on the night-market prompt.}
    \label{fig:cross-llm}
\end{figure}

Gemini achieves a slightly higher CLIPScore than DeepSeek ($0.220$ vs.\
$0.216$) but is slower ($374.8$ vs.\ $203.2$ seconds). Their aggregate VLM
preferences are near parity ($51.2\%$ vs.\ $48.9\%$): Gemini is favored for
alignment ($52.4\%$) and plausibility ($53.0\%$), while DeepSeek is favored for
completeness ($51.9\%$).
Figure~\ref{fig:cross-llm} presents a qualitative comparison across all four
language models for the same prompt.
These deployments show that the shared protocol accommodates multiple model
families. The near-parity comparison between Gemini and DeepSeek illustrates
that model choice can change runtime substantially without a corresponding
overall preference gain.

\subsection{Reusable Experience}
\paragraph{Library-size ablation.}
We construct a frozen library of 50 human-approved Layout and Content skills
from development prompts disjoint from GameSceneBench. We compare no skills,
a random subset of 20 skills, and the full library, with all other
settings unchanged. Table~\ref{tab:experience-ablation} reports the results.

\begin{table}[t]
\caption{Effect of experience-library size.}
\label{tab:experience-ablation}
\centering
\footnotesize
\textbf{(a) Absolute metrics by library size}\\[2pt]
\setlength{\tabcolsep}{6pt}
\begin{tabular}{lcc}
\toprule
Setting & CLIPScore $\uparrow$ & Runtime (s) $\downarrow$ \\
\midrule
\textsc{SceneAgent} (0 skills)
    & $0.211 \pm 0.031$ & $\mathbf{173.3 \pm 56.4}$ \\
\textsc{SceneAgent} (20 skills)
    & $0.214 \pm 0.032$ & $190.8 \pm 67.9$ \\
\textsc{SceneAgent} (50 skills)
    & $\mathbf{0.216 \pm 0.034}$ & $203.2 \pm 97.8$ \\
\bottomrule
\end{tabular}
\par\smallskip
\textbf{(b) Full-library preference vs. reduced library (\%)}\\[2pt]
\setlength{\tabcolsep}{6pt}
\begin{tabular}{lccc}
\toprule
Comparator & Align. $\uparrow$ & Plaus. $\uparrow$
    & Comp. $\uparrow$ \\
\midrule
0 skills  & $59.7$ & $46.2$ & $63.8$ \\
20 skills & $55.6$ & $50.5$ & $50.0$ \\
\bottomrule
\end{tabular}

\end{table}

The full library improves alignment and completeness over the no-skill
variant, with preference scores of $59.7\%$ and $63.8\%$, respectively. Spatial
plausibility is $46.2\%$, giving an aggregate preference of $56.6\%$ across
the three dimensions. Thus, the clearest benefit is selecting and organizing
prompt-relevant content. Additional guidance does not uniformly improve
geometry. CLIPScore increases from $0.211$ to $0.214$ to $0.216$ as the library
grows from 0 to 20 to 50 skills. The full library's aggregate preference over
20 skills is $52.1\%$, with similar plausibility and completeness, indicating
modest additional gains in this comparison. Runtime increases from $173.3$
to $203.2$ seconds. This ablation measures access to reusable experience.
It does not isolate factorization from alternative memory representations.

\begin{figure}[t]
    \centering
    \includegraphics[width=\columnwidth]{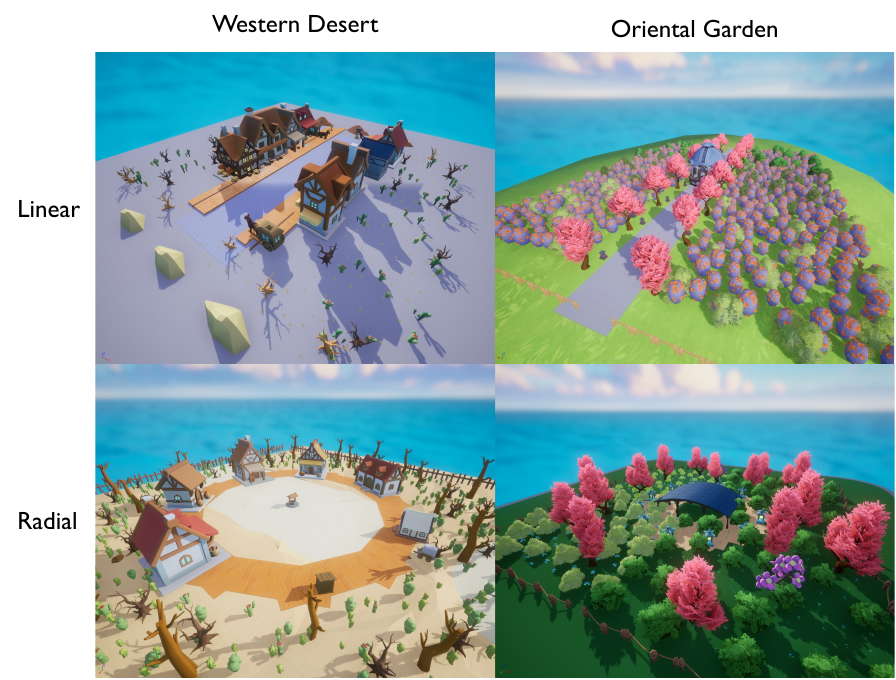}
    \caption{Composition of Layout and Content skills. Rows fix the Layout skill, while columns fix the Content skill.}
    \label{fig:skill-composition}
\end{figure}

\paragraph{Layout--Content composition.}
Figure~\ref{fig:skill-composition} crosses representative Layout skills
(rows) with Content skills (columns). Layout controls spatial organization,
whereas Content controls semantic and stylistic realization. The examples illustrate recombination at inference time without storing each
pairing separately. They illustrate the representation's compositional use.

\begin{figure}[t]
    \centering
    \includegraphics[width=\columnwidth]{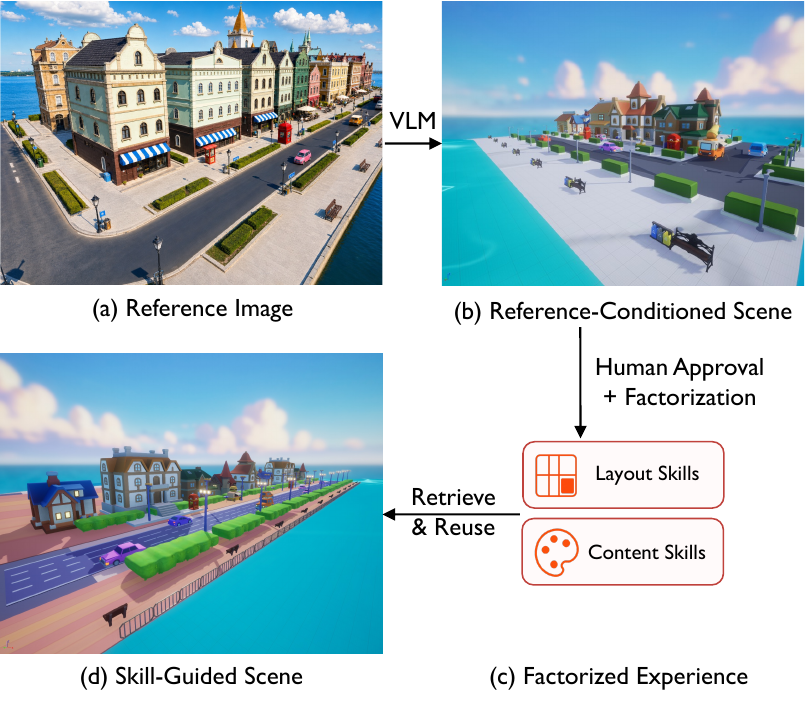}
    \caption{Visual experience acquisition and reuse. A reference image guides scene construction, and the human-approved trace is factorized into Layout and Content skills for reuse.}
    \label{fig:visual-experience}
\end{figure}

\paragraph{Visual experience acquisition.}
Figure~\ref{fig:visual-experience} illustrates how visual examples extend the
same experience library. Gemini 3.5 Flash~\cite{gemini35flash} extracts
semantic, spatial, and construction cues from a reference image for the
Planner--Builder pipeline. 
After human approval, the trace is factorized into Layout and Content skills through the standard update procedure. Later scenes reuse these skills without the source image.

\subsection{Single- versus Multi-Agent Construction}
We compare the full multi-agent \textsc{SceneAgent} with a monolithic
single-agent variant on all 69 GameSceneBench prompts. The single-agent
variant receives the original prompt and the complete construction-tool
interface, and follows a plan-then-construct workflow specified in its system
prompt. Unlike the full system, however, planning and construction are
performed by the same agent within a single persistent interaction history.
It does not instantiate separate Orchestrator, Planner, or Builder agents,
nor does it use structured inter-agent handoffs or fresh task-local contexts.

Both variants use the same language model backbone, Engine~A asset catalog,
retrieval service, construction tools, rendering protocol, and evaluation
procedure. We do not enforce an identical number of model or tool calls,
because these calls are determined by the respective execution protocols.
Runtime therefore reflects the end-to-end cost of each design. Each variant
is run three times on every benchmark prompt. Table~\ref{tab:agent-ablation}
reports the results.

\begin{table}[t]
\caption{Single- versus multi-agent construction on GameSceneBench. Results report mean $\pm$ standard deviation and tie-adjusted VLM preferences.}
\label{tab:agent-ablation}
\centering
\footnotesize
\setlength{\tabcolsep}{5pt}
\begin{tabular}{lcc}
\toprule
Metric & Single agent & Multi-agent \\
\midrule
CLIPScore $\uparrow$
    & $0.210 \pm 0.032$ & $\mathbf{0.216 \pm 0.034}$ \\
Runtime (s) $\downarrow$
    & $\mathbf{39.3 \pm 24.2}$ & $203.2 \pm 97.8$ \\
\midrule
VLM-Align. $\uparrow$ & $42.5$ & $\mathbf{57.5}$ \\
VLM-Plaus. $\uparrow$ & $49.0$ & $\mathbf{51.0}$ \\
VLM-Comp. $\uparrow$  & $37.7$ & $\mathbf{62.4}$ \\
\bottomrule
\end{tabular}

\end{table}

The multi-agent system improves CLIPScore from $0.210$ to $0.216$. In the
three VLM dimensions, it obtains tie-adjusted preference scores of $57.5\%$
for alignment, $51.0\%$ for spatial plausibility, and $62.4\%$ for scene
completeness. Aggregated across dimensions, it wins $54.6\%$ of cases, ties
in $4.8\%$, and loses $40.6\%$, corresponding to an overall preference score
of $57.0\%$.
The largest preference gain is in completeness, while plausibility is close
to parity. Average runtime increases from $39.3$ to $203.2$ seconds with the
additional planning, construction, and coordination stages. The result
supports the integrated protocol for more complete construction, with a
substantial quality--efficiency trade-off.

The ablations consistently favor coverage of requested content, while
spatial plausibility remains a challenge. The single-agent comparison tests
decomposition, context management, and handoffs jointly. Attributing gains
to individual mechanisms requires separate controls.

\section{Conclusion}
\textsc{SceneAgent} extends portability from scene output to the construction
process and its reusable experience. Canonical SceneContext state connects
fresh Builder sessions through structured handoffs, while factored Layout
and Content skills preserve design guidance across generations. Backend
adapters, asset configuration, and capability wrappers localize engine-specific
implementation. On GameSceneBench, human and VLM comparisons favor
\textsc{SceneAgent} over five baselines under a shared evaluation protocol.
Ablations show the clearest benefits in alignment and completeness, and
second-engine deployment demonstrates reuse of the shared protocol and
skills. Persistent world state and composable experience provide a practical basis
for portable scene construction.

\paragraph{Limitations.}
Porting still requires backend integration and compatible assets. The
cross-engine study uses different catalogs and establishes qualitative
retargeting rather than matched quantitative quality. Fresh Builder sessions
avoid replaying raw histories, but do not guarantee constant context size as
scenes grow. Multi-agent orchestration adds runtime, and human approval limits update
scalability. The ablations test the integrated protocol and experience
availability, leaving individual mechanisms and alternative memory
representations for further study. Finally, rendered-view
preferences and geometry checks do not establish physical stability,
navigability, or gameplay success. These limits motivate more efficient
orchestration, geometry-aware refinement, and broader validation of
experience transfer.
\section*{LLM Usage Statement}
The research uses DeepSeek-V4-Flash for scene construction, with Gemini 3.5
Flash, Hy3, and Qwen3.6-35B-A3B in cross-model experiments; Gemini 3.5 Flash
also supports visual experience acquisition. GPT-4o (\texttt{gpt-4o-2024-11-20})
is used for the pairwise scene evaluation described in the paper.
OpenAI GPT and Codex assisted with editing and refining author-produced manuscript
content. The authors remain responsible for all scientific claims,
experimental results, citations, and the final manuscript.

{\small
\bibliographystyle{ieeenat_fullname}
\bibliography{main}
}

\clearpage
\appendix
\maketitlesupplementary
\def\SceneAgentCombined{}
This supplementary material provides the complete benchmark specification,
implementation and evaluation details, complete outcome distributions,
cross-engine export examples, representative factored skills, and additional
qualitative results. It contains the following:
\begin{enumerate}
    \item The composition, category coverage, and complete prompt list of
    GameSceneBench.
    \item The construction and agent-control interfaces, including the
    implementation of entity-first asset retrieval.
    \item The rendering, runtime, and automated-evaluation protocols,
    including the pairwise VLM evaluation protocol.
    \item The human-evaluation protocol and study interface.
    \item The common-backend adaptation and runtime accounting for the
    evaluated baselines.
    \item Complete win, loss, and tie distributions for the main comparison.
    \item Cross-engine serialization examples and representative Layout,
    Content, and thin-recipe records.
    \item Additional qualitative results from Engines~A and B.
\end{enumerate}

\section{GameSceneBench Specification}
\label{sec:supp-benchmark}

\subsection{Benchmark Composition}
GameSceneBench is a text-driven benchmark for complete 3D game scene
generation. It contains 69 user prompts, comprising 56 outdoor scenes and 13
indoor scenes. Each entry contains a unique scene identifier, a display name,
an indoor--outdoor subset label, and the prompt supplied to every evaluated
method. The prompts range from short, open-ended requests to detailed
descriptions that specify required objects and spatial relations. The benchmark
therefore tests both broad scene completion and instruction following under
compositional constraints.

The outdoor and indoor subsets cover the scene families summarized in
Table~\ref{tab:supp-categories}. These categories summarize the types of scenes covered by the benchmark. They do not prescribe a unique layout or visual appearance for each prompt.

Each method receives the same input prompt and produces a complete 3D scene using the common asset catalog, rendering pipeline, and standardized camera protocol. Every method is run independently three times for each prompt to account for generation variability. Because the benchmark admits multiple valid scene realizations and does not define a unique reference scene, the evaluation measures text--scene alignment, generation efficiency, spatial plausibility, and scene completeness rather than reconstruction error against a single fixed ground truth.

\subsection{Complete Benchmark Prompts}
The following list provides the complete set of 69 prompts used in GameSceneBench. These prompts cover a diverse range of scene types, visual themes, spatial organizations, and content requirements, and are used consistently across all methods and evaluation settings. Stable scene identifiers are shown in monospace to make each prompt directly traceable and comparable with generation outputs, rendering results, and evaluation records. A machine-readable transcription of this list is included as \texttt{gamescenebench.json} in the supplementary archive.

\begin{table}[t]
\caption{Category coverage of GameSceneBench.}
\label{tab:supp-categories}
\centering
\footnotesize
\setlength{\tabcolsep}{5pt}
\begin{tabular}{p{0.2\columnwidth}p{0.68\columnwidth}}
\toprule
Subset & Scene families \\
\midrule
Outdoor (56) & Natural landscapes, settlements, sports and recreational
spaces, historical and stylized environments, and urban infrastructure. \\
Indoor (13) & Residential, workplace, commercial, hospitality,
entertainment, and utility spaces. \\
\bottomrule
\end{tabular}

\end{table}

\paragraph{Outdoor scenes (56).}
{\footnotesize
\begin{enumerate}
    \item \texttt{park\_square}: Generate an urban park square scene.
    \item \texttt{mongolia\_grassland}: Generate an Inner Mongolian grassland scene.
    \item \texttt{rural\_village}: Generate a rural village scene.
    \item \texttt{winter\_town}: Generate a winter town scene.
    \item \texttt{ice\_wonderland}: Generate an ice-and-snow wonderland with a scenic promenade, a giant ice slide, a snowflake-shaped Ferris wheel, and lively interactive winter attractions that convey the distinctive atmosphere of an ice city.
    \item \texttt{primitive\_tribe}: Generate a primitive tribal settlement scene.
    \item \texttt{egypt\_adventure}: Generate an ancient Egyptian expedition scene with a wooden rail track and mine cart leading to a cave, as well as expedition tents and wooden crates.
    \item \texttt{chinese\_garden}: Generate a classical Chinese garden with traditional buildings, a pond, a small pavilion, and covered corridors.
    \item \texttt{epic\_valley}: Generate a valley flanked by steep mountain ranges covered with fir forests. A winding river should cross the valley floor and form pebble banks and shallows, with dense vegetation and rocks in the foreground.
    \item \texttt{summer\_field}: Generate a summer field scene.
    \item \texttt{venice}: Generate a Venetian canal-city scene.
    \item \texttt{school\_track}: Generate a school athletic field surrounded by classroom buildings and lush trees.
    \item \texttt{mt\_fuji}: Generate a scene at the foot of snow-covered Mount Fuji, where a vintage tram travels along a quiet Japanese street lined with blooming pink cherry trees, creating a beautiful and soothing atmosphere.
    \item \texttt{flower\_garden}: Generate a small flower-garden scene.
    \item \texttt{beach\_resort}: Generate a resort beach scene.
    \item \texttt{volleyball\_court}: Generate a beach volleyball scene.
    \item \texttt{japanese\_onsen}: Generate an outdoor Japanese hot-spring scene.
    \item \texttt{pine\_forest}: Generate a pine forest scene.
    \item \texttt{zen\_garden}: Generate a dry-landscape Zen garden.
    \item \texttt{baseball\_field}: Generate a baseball field scene.
    \item \texttt{sakura\_grove}: Generate a cherry-blossom grove.
    \item \texttt{european\_garden}: Generate a European-themed garden.
    \item \texttt{pond\_stream}: Generate a pond-and-stream scene.
    \item \texttt{desert\_gobi}: Generate a desert and Gobi landscape.
    \item \texttt{desert\_oasis}: Generate a desert oasis surrounded by sparse cacti of different species and several dead trees.
    \item \texttt{desert\_camp}: Generate a desert-oasis campsite surrounded by scattered dead trees. Place a clear circular pool at the center, with dense flowers, shrubs, and several tall palm trees around it. Include a stone-built water point and several tents beside the pool.
    \item \texttt{swimming\_pool}: Generate an outdoor swimming pool with white lounge chairs and sun umbrellas along the poolside, conveying a relaxed vacation atmosphere.
    \item \texttt{creek\_canyon}: Generate a stream-canyon scene.
    \item \texttt{tropical\_beach}: Generate a tropical palm beach.
    \item \texttt{tropical\_resort}: Generate a tropical seaside resort.
    \item \texttt{tomb\_raider}: Generate a tomb-exploration adventure scene.
    \item \texttt{bamboo\_forest}: Generate a bamboo forest scene.
    \item \texttt{basketball\_court}: Generate a regulation basketball court.
    \item \texttt{tennis\_court}: Generate a tennis court scene.
    \item \texttt{soccer\_field}: Generate a regulation soccer field.
    \item \texttt{snow\_camp}: Generate a snowy campsite enclosed by dense snow-covered coniferous forest. The central clearing should have level snow, several tents, and a wind-sheltered fire pit built from stones. Place several wooden stakes around the campsite as boundary markers.
    \item \texttt{chinese\_temple}: Generate a traditional Chinese temple scene.
    \item \texttt{military\_base}: Generate a military base scene.
    \item \texttt{military\_march}: Generate a scene of a Chinese military unit marching through the countryside.
    \item \texttt{villa\_town}: Generate a villa town with several neatly arranged modern villas along level streets, each with a small garden. Include a circular fountain plaza at the center, and enclose the scene with dense trees and orderly hedges.
    \item \texttt{ancient\_city}: Generate a scene depicting an assault on a fortified city in ancient China.
    \item \texttt{ancient\_ruins}: Generate ancient architectural ruins with broken walls, drifting sand, and overgrown weeds.
    \item \texttt{subway\_station}: Generate a subway station scene.
    \item \texttt{city\_street}: Generate an urban street scene with a roadway and ordinary buildings.
    \item \texttt{night\_market}: Generate a night-market food street.
    \item \texttt{wasteland\_defense}: Generate a post-apocalyptic wasteland defense position centered on a circular base enclosed by abandoned cars, barbed wire, and sandbags. Include a campfire and several worn tents inside the base.
    \item \texttt{airport}: Generate an international airport scene.
    \item \texttt{forest\_town}: Generate a forest town surrounded by dense woodland. At its center, stone-paved streets connect rows of wooden and brick houses. Include an open-air market in the central square with stalls displaying furs, herbs, and handicrafts, and place a fenced stable and a blacksmith shop at the edge of the settlement.
    \item \texttt{amusement\_park}: Generate an amusement park scene.
    \item \texttt{banana\_factory}: Generate a banana-processing factory.
    \item \texttt{tang\_dynasty}: Generate a prosperous Tang-dynasty Chang'an scene with a lively food street, an ornate palace, and lanterns filling the sky.
    \item \texttt{western\_cowboy}: Generate an American Western scene in which a train departs along a railroad bordered by sparse cacti of different species.
    \item \texttt{western\_town}: Build an American frontier-era cowboy town with wooden buildings lining both sides of a main street.
    \item \texttt{supermarket\_parking}: Generate a Sam's Club parking-lot scene.
    \item \texttt{marketplace}: Generate a marketplace scene.
    \item \texttt{graveyard}: Generate a graveyard scene.
\end{enumerate}
}

\paragraph{Indoor scenes (13).}
{\footnotesize
\begin{enumerate}
    \item \texttt{indoor\_cozy\_bedroom\_tv}: Generate a cozy double bedroom with the head of the bed against a wall, one nightstand and table lamp on each side, and a wall-mounted television directly opposite the foot of the bed.
    \item \texttt{indoor\_living\_room}: Generate a modern apartment living room.
    \item \texttt{indoor\_dining\_room}: Generate a family dining room with a rectangular dining table at the center, four chairs facing the table, a sideboard against a wall, and a pendant light directly above the table.
    \item \texttt{indoor\_home\_office}: Generate a home office with a desk against a wall, an office chair facing the desk, a bookcase against a wall, and a reading corner composed of a floor lamp and a small plant.
    \item \texttt{indoor\_kitchen}: Generate a kitchen.
    \item \texttt{indoor\_meeting\_room}: Generate a large meeting room.
    \item \texttt{indoor\_gym}: Generate the interior of a gym.
    \item \texttt{indoor\_fruit\_store}: Generate the interior of a fruit store.
    \item \texttt{indoor\_hotel\_lobby}: Generate a hotel lobby.
    \item \texttt{indoor\_laundry\_room}: Generate a laundry room.
    \item \texttt{indoor\_nursery}: Generate a nursery.
    \item \texttt{indoor\_gaming\_room}: Generate a gaming room.
    \item \texttt{indoor\_garage}: Generate a garage interior.
\end{enumerate}
}

\section{Agent Tool Interfaces}
\label{sec:supp-tools}

\subsection{Builder Construction Interface}
Builder agents operate on the canonical SceneContext through the shared
tool interface summarized in Table~\ref{tab:supp-builder-tools}. The interface
covers the complete construction loop: grounding assets, placing individual
objects, synthesizing parametric structures, populating regions, inspecting
the evolving scene, and making targeted corrections. It also provides a
consistent execution boundary through which Builders can query existing scene
state, validate intermediate results, and incrementally refine previously
constructed content without directly accessing backend-specific APIs. Tools
consume and return builder-coordinate quantities, while backend adapters handle
engine-specific coordinates, units, asset bindings, command execution, and
scene serialization. This separation allows the same Builder policy and
construction protocol to remain unchanged across supported backends.

\begin{table}[t]
\caption{Builder-visible construction tools. The interface is shared across
backends. Backend-specific configuration supplies compatible asset categories
and export conventions.}
\label{tab:supp-builder-tools}
\centering
\footnotesize
\setlength{\tabcolsep}{4pt}
\begin{tabular}{p{0.3\columnwidth}p{0.61\columnwidth}}
\toprule
Tool family & Tools \\
\midrule
Asset discovery and grounding &
\texttt{search\_modules}, \texttt{get\_module\_info},
\texttt{web\_search}, \texttt{view\_thumbnails},
\texttt{plan\_assets}, \texttt{get\_planned\_assets} \\
Direct and relation-aware placement &
\texttt{place\_module}, \texttt{place\_facing},
\texttt{place\_against\_wall}, \texttt{place\_on\_ceiling} \\
Parametric geometry and structures &
\texttt{create\_floor}, \texttt{build\_room},
\texttt{create\_line}, \texttt{create\_rect},
\texttt{create\_circle}, \texttt{create\_arc},
\texttt{build\_column}, \texttt{build\_board},
\texttt{build\_stairs}, \texttt{create\_shape} \\
Procedural layout and population &
\texttt{region\_define}, \texttt{get\_landscape\_data},
\texttt{surface\_sampler}, \texttt{transform\_points},
\texttt{remove\_points\_by\_region}, \texttt{batch\_place},
\texttt{pcg\_scatter} \\
Scene-state queries and handoff &
\texttt{get\_scene\_summary}, \texttt{update\_conventions},
\texttt{report\_subtask} \\
Targeted editing and rollback &
\texttt{undo\_last}, \texttt{clear\_scene},
\texttt{remove\_by\_type}, \texttt{remove\_by\_region},
\texttt{undo\_subtask} \\
Geometry verification &
\texttt{check\_bbox\_collision} \\
\bottomrule
\end{tabular}

\end{table}

\paragraph{Entity-First Asset Retrieval with \texttt{search\_modules}.}
\texttt{search\_modules} maps a natural-language asset request to a ranked
list of candidates from the active backend's asset catalog. Textual asset
cards are encoded offline and stored in an approximate nearest-neighbor index,
while each incoming query is encoded online and ranked by cosine similarity.
The retriever follows an entity-first objective: candidates should match the
requested object category before satisfying secondary attributes such as
style, material, condition, or scene context. Its training data combine
asset-grounded query synthesis with semantically similar hard negatives,
including deceptive negatives that match the requested attributes but
represent the wrong entity. A shared encoder is trained with a standard
InfoNCE objective. The returned top-$K$ candidates remain proposals rather
than final bindings. Agents may inspect their metadata and thumbnails before
selecting an asset.

\subsection{Agent-Control Interface}
The role-specific tools in Table~\ref{tab:supp-control-tools} delimit the
responsibilities of the Orchestrator, Planner, and Builders. The Orchestrator
does not place actors directly, and the Planner commits semantic subtasks
rather than modifying SceneContext. Final serialization is
kept outside individual Builder sessions.

\begin{table}[t]
\caption{Role-specific control interfaces. Area-budget estimation is enabled
for area-aware planning, and room-size estimation is enabled for indoor
planning.}
\label{tab:supp-control-tools}
\centering
\footnotesize
\setlength{\tabcolsep}{4pt}
\begin{tabular}{p{0.23\columnwidth}p{0.67\columnwidth}}
\toprule
Agent & Control tools \\
\midrule
Orchestrator & \texttt{load\_skill}, \texttt{plan\_scene},
\texttt{revise\_plan}, \texttt{build\_scene},
\texttt{refine\_scene}, \texttt{export\_scene} \\
Planner & \texttt{search\_modules}, \texttt{get\_module\_info},
\texttt{web\_search}, \texttt{view\_thumbnails},
\texttt{estimate\_area\_budget}, \texttt{estimate\_room\_size},
\texttt{commit\_subtasks} \\
Builder & \texttt{update\_conventions}, \texttt{report\_subtask}, plus the
construction tools in Table~\ref{tab:supp-builder-tools} \\
\bottomrule
\end{tabular}

\end{table}

The low-level \texttt{export\_json} operation is hidden from Builder sessions
and invoked through the Orchestrator's \texttt{export\_scene} interface. This
keeps construction backend-neutral and localizes final serialization to the
active backend adapter.

\FloatBarrier
\section{Rendering and Automated Evaluation}
\label{sec:supp-evaluation}

\subsection{Five-View Rendering Protocol}
For each generated scene, the camera rig is centered on the bounding box that
encloses all valid actors. Its distance is adjusted automatically according to
the scene extent. We render four perspective views distributed around the scene
at an elevation angle of approximately $30^\circ$, together with one top-down
view. The perspective views capture the scene from complementary directions,
whereas the top-down view exposes the global layout, object distribution, and
spatial relations. All views use a $60^\circ$ field of view and a resolution of
$1920\times1080$.

The same camera-placement and rendering procedure is used for all methods. This
controlled setup reduces variation from presentation and helps isolate
differences in planning and scene construction.

\subsection{CLIPScore and Runtime}
We compute CLIPScore~\cite{hessel2021clipscore} using CLIP
ViT-L/14~\cite{radford2021learning}. For each generated scene, we compute the
score between the input prompt and each of the five rendered views and average
the five per-view scores. We then average over the three independent runs for
each prompt when aggregating benchmark results.

Runtime is wall-clock time from prompt submission to completion of the scene
file. It includes planning and construction and excludes offline rendering, VLM
evaluation, and human evaluation. Runtime is reported in seconds and averaged
over the three independent runs.

\subsection{Pairwise VLM Evaluation}
The VLM evaluation considers three complementary dimensions:
\begin{itemize}
    \item \textbf{Text--scene alignment}: whether the scene satisfies the
    objects, style, spatial relations, and other constraints in the prompt.
    \item \textbf{Spatial plausibility}: object placement, support, scale,
    accessibility, and unreasonable overlap.
    \item \textbf{Scene completeness}: whether the scene contains sufficient,
    nonredundant supporting content to form a coherent environment.
\end{itemize}

For each method pair, prompt, and run, the five views of both scenes are
letterboxed into an anonymous $2\times5$ composite. The assignment of methods
to the top and bottom rows is randomized. The VLM answers a separate pairwise
question for each dimension with three possible outcomes: top, bottom, or tie.
We repeat each comparison with independently randomized positions. The main
experiments use \texttt{gpt-4o-2024-11-20} for this evaluation.

For a method $A$, the reported tie-adjusted preference is
\begin{equation}
    P_A = \frac{W_A + 0.5T}{W_A + L_A + T}\times100\%,
\end{equation}
where $W_A$, $L_A$, and $T$ denote the numbers of wins, losses, and ties for
$A$, respectively. Thus, $50\%$ indicates parity.

\section{Human Evaluation Protocol}
\label{sec:supp-human}

\paragraph{Study oversight and consent.}
The human evaluation consisted of preference judgments between rendered
synthetic scenes. No formal institutional ethics review was conducted because
the study was limited to scene-preference evaluation. All participants were
informed about the study and provided informed consent before taking part.

We conducted a pairwise human preference study using the same anonymous
$2\times5$ composites as in the VLM evaluation. Participants saw the input
prompt and the two sets of rendered views, with method identities hidden and
the top--bottom order randomized. For each comparison, they selected the scene
with better overall plausibility and consistency with the input text, or
indicated no preference. No preference is counted as half a win when computing
the tie-adjusted score.

\begin{figure}[t]
    \centering
    \includegraphics[width=\columnwidth]{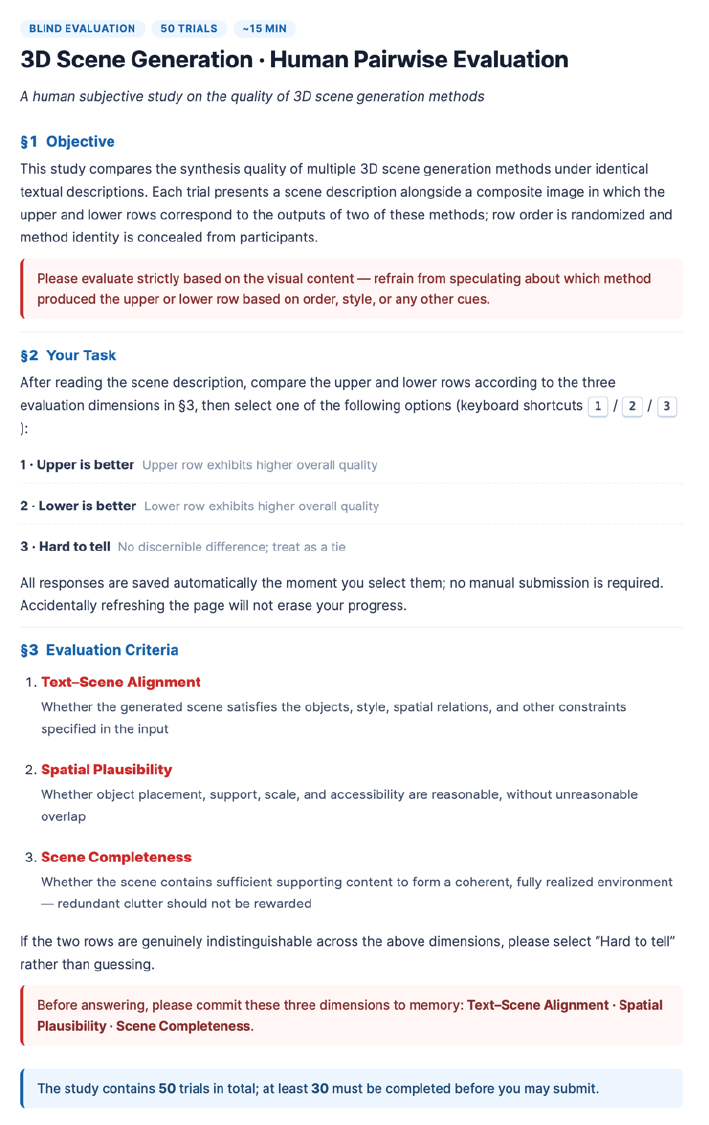}
    \caption{Instructions presented to participants before the human pairwise
    evaluation. The page describes the blind protocol, response options,
    evaluation dimensions, and minimum completion requirement.}
    \label{fig:user-study-instruction}
\end{figure}

The study interface offered 50 trials and required at least 30 completed
comparisons, as shown in Figure~\ref{fig:user-study-instruction}.
We recruited 65 volunteers. 14 participants were excluded because they
did not complete the required number of comparisons, leaving 51 participants
and 1,860 retained judgments. The judgments were approximately
balanced across comparisons between \textsc{SceneAgent} and the five baselines.

\begin{figure*}[t]
    \centering
    \includegraphics[width=0.98\textwidth]{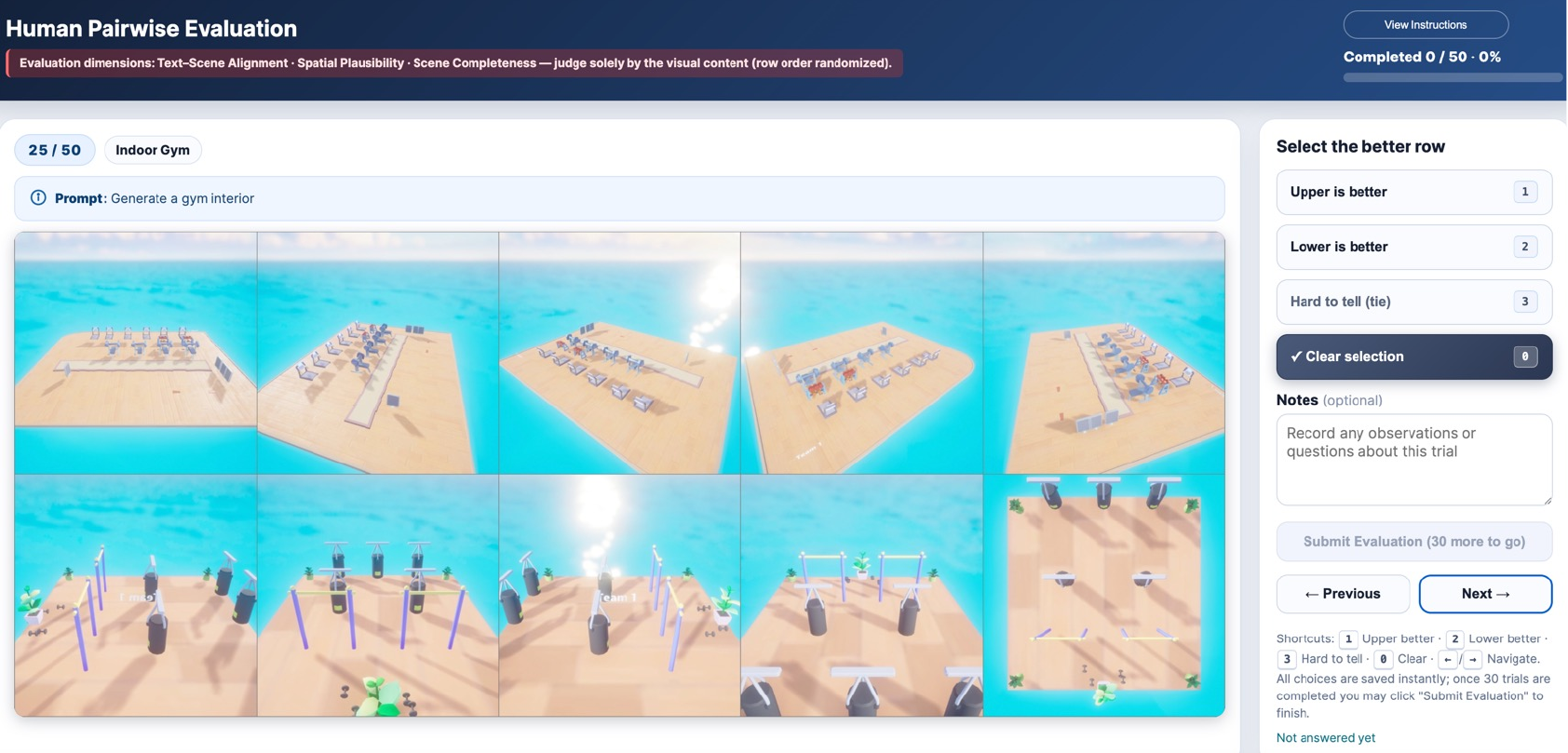}
    \caption{Human pairwise-evaluation interface. Each trial presents the scene
    prompt and an anonymous, randomly ordered $2\times5$ composite containing
    five views per method. Participants select the better row or a tie, may
    record optional notes, and navigate through the assigned trials.}
    \label{fig:user-study-interface}
\end{figure*}

The instruction screen shown before evaluation is reproduced in
Figure~\ref{fig:user-study-instruction}. It explains the blind comparison
setting, response options, three evaluation dimensions, keyboard shortcuts,
and completion requirement. Figure~\ref{fig:user-study-interface} shows the
trial interface. The prompt and anonymous $2\times5$ composite occupy the main
panel, while the response controls, optional notes, trial navigation, and
progress indicator remain visible in a fixed side panel.

\FloatBarrier
\section{Baseline Adaptation \& Runtime Accounting}
\label{sec:supp-baselines}

\subsection{Baseline-Specific Adaptations}
We compare \textsc{SceneAgent} against five representative baselines spanning
distinct paradigms of automatic 3D scene generation. AutoUE~\cite{yin2026autoue}
prompts an LLM to author a procedural scene graph executed by a game-engine
runtime. Holodeck~\cite{yang2024holodeck} synthesizes indoor layouts through
retrieval-guided constraint solving. ImperativeScene~\cite{gumin2025procedural}
expresses scene layouts as LLM-generated imperative programs.
Reason-3D~\cite{berdoz2026text} pairs LLM-driven spatial reasoning with
embedding-based asset retrieval. SceneWeaver~\cite{yang2026sceneweaver}
integrates LLM-based scene planning with a procedural asset-generation engine.

These methods were originally developed around mutually incompatible asset
ecosystems and rendering backends, many of which rely on proprietary,
online-only, or platform-specific components. Their original pipelines
therefore cannot be deployed uniformly and reproducibly in a single evaluation
environment. To enable a fair comparison, we standardize three infrastructure
layers across all five baselines:
\begin{itemize}
    \item \textbf{Asset retrieval}: retrieval is redirected to a shared 3D
    asset repository accessed through a common text-based retrieval service.
    Each method retains its own search prompt or query formulation.
    \item \textbf{Execution and rendering}: layout execution and rendering are
    routed through the shared Engine~A pipeline and camera protocol.
    \item \textbf{Language model}: every method uses DeepSeek-V4-Flash
    ~\cite{xu2026deepseek} with a context window of up to 204,800 tokens.
\end{itemize}

The adaptations retain each method's core scene-generation procedure,
including its graph structure, constraint formulation, program templates,
or reasoning workflow, within the common execution environment. All five
baselines and \textsc{SceneAgent} use the same asset repository, retrieval
service, rendering pipeline, and language model. This controls major sources
of infrastructure variation and compares the adapted generation procedures.
It does not reproduce each original system's full native capabilities.
Results should be interpreted within the common-backend setting.

\subsection{AutoUE Adaptation}
AutoUE~\cite{yin2026autoue} is designed for end-to-end game generation in
Unreal Engine. For the common-backend comparison, we evaluate its scene-layout
stage. The adaptation retains layout generation, asset grounding, and scene
instantiation, while excluding Unreal-Engine-specific gameplay, interaction,
and testing stages. This condition is denoted AutoUE$^\dagger$ in the result
tables.

\subsection{Runtime Boundary}
The same runtime boundary is applied to every evaluated method: measurement
begins when the prompt is submitted and ends when scene generation completes.
Planning, model interaction, retrieval, tool invocation, and construction are
included when performed by the method. Offline rendering, VLM scoring, and
human evaluation are excluded. This measures end-to-end generation cost while
avoiding evaluation-time overhead.

For \textsc{SceneAgent}, the Orchestrator is limited to 30 interaction rounds,
the Planner to 50 rounds and 40 tool calls, and each Builder to 40 rounds. The
Planner produces 5--10 subtasks for outdoor scenes and 3--6 subtasks for indoor
scenes. The Orchestrator may invoke at most three automatic refinement rounds.
The experience library is frozen before evaluation and is not updated using any
GameSceneBench generation or feedback.

\section{Complete Outcome Distributions and Cross-Engine Details}
\label{sec:supp-results}

\subsection{Main Comparison on GameSceneBench}
The main paper reports the aggregate CLIPScore, runtime, and tie-adjusted
preferences. Table~\ref{tab:supp-main-outcomes} instead provides the underlying
three-way outcome distributions, preserving how frequently \textsc{SceneAgent}
wins outright, the baseline wins outright, or a comparison is judged a tie.
The human-study panel also gives the exact number of retained judgments for
each baseline comparison. These counts sum to 1,860.

\begin{table*}[t]
\caption{Complete pairwise outcome distributions for the main comparison.
VLM entries report mean $\pm$ standard deviation. ``Overall'' aggregates the
three VLM dimensions. Human percentages are computed over the retained
judgments for each comparator and may not sum to exactly $100\%$ after
rounding.}
\label{tab:supp-main-outcomes}
\centering
\footnotesize
\textbf{(a) Raw VLM outcomes by comparator and dimension (\%)}\\[2pt]
\setlength{\tabcolsep}{5pt}
\begin{tabular}{llccc}
\toprule
Comparator & Dimension & \textsc{SceneAgent} win & Baseline win & Tie \\
\midrule
AutoUE$^\dagger$ & Overall & $85.4 \pm 1.6$ & $12.1 \pm 2.1$ & $2.6 \pm 1.5$ \\
 & Alignment & $94.2 \pm 2.9$ & $5.8 \pm 2.9$ & $0.0 \pm 0.0$ \\
 & Plausibility & $70.5 \pm 2.2$ & $21.7 \pm 3.8$ & $7.7 \pm 4.4$ \\
 & Completeness & $91.3 \pm 0.0$ & $8.7 \pm 0.0$ & $0.0 \pm 0.0$ \\
\midrule
Holodeck & Overall & $76.2 \pm 3.6$ & $21.1 \pm 3.4$ & $2.7 \pm 0.8$ \\
 & Alignment & $81.2 \pm 2.9$ & $18.4 \pm 2.2$ & $0.5 \pm 0.8$ \\
 & Plausibility & $70.5 \pm 10.2$ & $22.2 \pm 9.3$ & $7.2 \pm 2.9$ \\
 & Completeness & $76.8 \pm 3.8$ & $22.7 \pm 4.2$ & $0.5 \pm 0.8$ \\
\midrule
ImperativeScene & Overall & $71.0 \pm 2.7$ & $26.1 \pm 2.2$ & $2.9 \pm 0.5$ \\
 & Alignment & $77.3 \pm 3.3$ & $22.7 \pm 3.3$ & $0.0 \pm 0.0$ \\
 & Plausibility & $60.9 \pm 1.4$ & $30.4 \pm 0.0$ & $8.7 \pm 1.4$ \\
 & Completeness & $74.9 \pm 4.7$ & $25.1 \pm 4.7$ & $0.0 \pm 0.0$ \\
\midrule
Reason-3D & Overall & $72.8 \pm 1.6$ & $24.5 \pm 0.2$ & $2.7 \pm 1.4$ \\
 & Alignment & $77.8 \pm 3.0$ & $21.3 \pm 3.0$ & $1.0 \pm 1.7$ \\
 & Plausibility & $58.0 \pm 5.0$ & $34.8 \pm 2.5$ & $7.2 \pm 2.5$ \\
 & Completeness & $82.6 \pm 2.5$ & $17.4 \pm 2.5$ & $0.0 \pm 0.0$ \\
\midrule
SceneWeaver & Overall & $72.3 \pm 2.3$ & $22.7 \pm 1.3$ & $5.0 \pm 2.2$ \\
 & Alignment & $82.5 \pm 1.3$ & $15.5 \pm 1.6$ & $1.9 \pm 0.8$ \\
 & Plausibility & $58.3 \pm 4.8$ & $29.6 \pm 0.7$ & $12.1 \pm 4.4$ \\
 & Completeness & $76.2 \pm 2.1$ & $22.8 \pm 3.0$ & $1.0 \pm 1.7$ \\
\bottomrule
\end{tabular}
\par\smallskip
\textbf{(b) Raw human-study outcomes by comparator (\%)}\\[2pt]
\setlength{\tabcolsep}{6pt}
\begin{tabular}{lccccc}
\toprule
Comparator & \textsc{SceneAgent} win & Baseline win & Tie & Judgments & Qualified sessions \\
\midrule
AutoUE$^\dagger$ & $83.4$ & $4.8$ & $11.9$ & $379$ & $51$ \\
Holodeck & $78.5$ & $8.2$ & $13.3$ & $368$ & $51$ \\
ImperativeScene & $71.8$ & $15.4$ & $12.8$ & $383$ & $51$ \\
Reason-3D & $75.0$ & $12.6$ & $12.4$ & $364$ & $51$ \\
SceneWeaver & $76.0$ & $10.7$ & $13.4$ & $366$ & $51$ \\
\bottomrule
\end{tabular}

\end{table*}

\subsection{Cross-Engine Portability}
The cross-engine study runs all 69 prompts in Engine~B with the same language
model, agent protocol, SceneContext, core construction interface, skill
library, and generation budget. The adapter, asset catalog and database,
and capability wrappers are backend-specific. Because the catalogs differ,
the study demonstrates deployment through this boundary rather than a
matched quantitative comparison of engine quality. Here, abridged export
payloads illustrate how canonical state is serialized for each backend.

\paragraph{Backend export payloads.}
The canonical SceneContext is converted into the native import schema of
each target engine. Listings~\ref{lst:supp-engine-a-export}
and~\ref{lst:supp-engine-b-export} show abridged examples of the final files
delivered to Engines~A and B, respectively. Only one representative actor is
shown in each listing. Thus, the listings
illustrate the serialization boundary rather than reproducing the full scenes.

Engine~A stores the actor array directly. Asset identifiers are strings,
positions use the engine's centimeter convention, and each actor carries a
display name, an instance name, collision and mode flags, an instance index,
a quaternion rotation, scale, and optional color slots. For compactness,
floating-point values are rounded in the excerpt.

\begin{samepage}
\begin{lstlisting}[
  caption={Abridged Engine~A import payload.},
  label={lst:supp-engine-a-export},
  numbers=none,
  basicstyle=\scriptsize\ttfamily,
  breaklines=true,
  breakatwhitespace=true,
  columns=fullflexible,
  keepspaces=true,
  showstringspaces=false,
  xleftmargin=0pt
]
[
  {"actors": [{
    "typeId": "20009245",
    "name": "Simple Residential Stone Floor 01",
    "gname": "asset0", "c": 1,
    "pos": [0.0, 0.0, -5.0],
    "rot": [0.0, 0.0, 0.0, 1.0],
    "sca": [27.550, 26.732, 0.080],
    "col": [], "id": 0, "m": 1
  }]}
]
\end{lstlisting}
\end{samepage}

Engine~B wraps scene generation in a \texttt{PCGGenerate} command. Its
\texttt{Param.Info} field contains a serialized JSON object with the actor,
group, and relation records. Asset identifiers are numeric, positions are
converted to meters, and transform components are serialized as fixed-precision
strings. The per-export \texttt{AIID} is generated at runtime. The listing
separates the decoded \texttt{Info} payload solely to make the nested schema
readable.

\begin{samepage}
\begin{lstlisting}[
  caption={Abridged Engine~B import payload.},
  label={lst:supp-engine-b-export},
  numbers=none,
  basicstyle=\scriptsize\ttfamily,
  breaklines=true,
  breakatwhitespace=true,
  columns=fullflexible,
  keepspaces=true,
  showstringspaces=false,
  xleftmargin=0pt
]
[
  {"Type": "PCGGenerate",
   "Param": {
     "AIID": "<generated-id>",
     "bNeedGroup": false,
     "Info": "<serialized payload below>"
   }}
]

Decoded Param.Info:
{
  "actors": [{
    "typeId": 1101002001004002,
    "pos": ["0.000", "0.000", "-0.025"],
    "sca": ["40.000", "40.000", "0.017"],
    "rot": ["0.000", "0.000", "0.000", "1.000"],
    "c": 0, "col": [[235, 240, 255]]
  }],
  "group": {},
  "relation": []
}
\end{lstlisting}
\end{samepage}

\section{Reusable Experience Update}
\label{sec:supp-experience-update}

\subsection{Human-Led Quality Gate}
The main paper describes the human-led quality gate and the roles of automatic
diagnostics, schema validation, and deduplication. Algorithm~\ref{alg:supp-experience-update}
provides the corresponding procedural specification.

\begin{algorithm}[t]
\caption{Human-Led Quality-Gated Experience Update}
\label{alg:supp-experience-update}
\begin{algorithmic}[1]
\STATE Initialize Layout library $\mathcal{L}$, Content library $\mathcal{C}$,
and recipe library $\mathcal{R}$
\FOR{each generation trace $\tau$}
    \STATE Collect SceneContext, renderings, and automatic diagnostics
    \STATE Present $\tau$ and its diagnostics for human review
    \IF{the reviewer approves $\tau$}
        \STATE Extract Layout candidates $\hat{L}$, Content candidates
        $\hat{C}$, and recipe $r$
        \STATE Reject candidates that violate the transferable-skill schemas
        \STATE Deduplicate and update $\mathcal{L}$ and $\mathcal{C}$; insert
        $r$ into $\mathcal{R}$
    \ELSE
        \STATE Store $\tau$ and reviewer feedback as failure evidence
    \ENDIF
\ENDFOR
\end{algorithmic}
\end{algorithm}

\subsection{Representative Factored-Skill Examples}
Listings~\ref{lst:supp-layout-skill}--\ref{lst:supp-thin-recipe} show an
abridged, mutually compatible triplet from the experience library. The Layout
skill contains only reusable spatial organization and geometric constraints.
The Content skill specifies semantic roles, stylistic attributes, and
substitution rules without backend-specific asset bindings. The thin
recipe references both skills while retaining only requirements specific to
the target scene.

\newpage
\begin{samepage}
\begin{lstlisting}[
  caption={Representative Layout skill containing backend- and
  asset-independent linear organization.},
  label={lst:supp-layout-skill},
  numbers=none,
  basicstyle=\scriptsize\ttfamily,
  breaklines=true,
  breakatwhitespace=true,
  columns=fullflexible,
  keepspaces=true,
  showstringspaces=false,
  xleftmargin=0pt
]
type: linear_guide
category: layout
name: Linear Guided Layout
when_to_use: >
  Scenes organized around one dominant route with short
  branches that quickly reconnect to the main path.

topology:
  structure: linear
  anchors: [entrance, terminal_focus]
  main_axis: shared_by_all_route_segments

spatial_rules:
- Connect the entrance and terminal focus with a continuous route.
- Arrange major regions along both sides of the route.
- Preserve an unobstructed circulation corridor.
- Orient flanking structures toward the route.
- Frame the scene without blocking traversal.

constraints:
- Derive all route segments from one consistent axis.
- Validate footprints before placing large structures.
- Align directional elements with the route.
\end{lstlisting}
\end{samepage}

\begin{samepage}
\begin{lstlisting}[
  caption={Representative Content skill containing semantic and stylistic
  guidance without concrete asset identifiers.},
  label={lst:supp-content-skill},
  numbers=none,
  basicstyle=\scriptsize\ttfamily,
  breaklines=true,
  breakatwhitespace=true,
  columns=fullflexible,
  keepspaces=true,
  showstringspaces=false,
  xleftmargin=0pt
]
type: night_market_chinese_palette
category: content
name: East Asian Night-Market Content
when_to_use: >
  Lively night markets and pedestrian food streets with
  warm illumination and dense everyday details.

semantic_roles:
  vendors: [food stall, snack booth, market cart, canopy]
  lighting: [red lantern, string lights, neon sign]
  furniture: [small table, chair, bench, stool]
  supporting_props: [crate, signboard, waste bin,
                     utility box]

style:
  palette:
    base: [dark gray, asphalt black, warm wood]
    lighting: [red, orange, yellow]
    accents: [saturated red, blue, green]
  appearance: stylized or low-poly
  atmosphere: lively, warm, dense, and informal

substitution_rules:
- Replace a missing food stall with a canopy-and-counter assembly.
- Replace neon signage with illuminated hanging signs.
- Vary vendor and furniture types to avoid repetition.
\end{lstlisting}
\end{samepage}

\begin{samepage}
\begin{lstlisting}[
  caption={Representative thin recipe containing scene-specific residuals
  for the selected Layout and Content skills.},
  label={lst:supp-thin-recipe},
  numbers=none,
  basicstyle=\scriptsize\ttfamily,
  breaklines=true,
  breakatwhitespace=true,
  columns=fullflexible,
  keepspaces=true,
  showstringspaces=false,
  xleftmargin=0pt
]
type: night_market
category: recipe
name: Lively Night-Market Street

composed_of:
  layout: linear_guide
  content: night_market_chinese_palette

residual_requirements:
- Place two staggered rows of vendors along a pedestrian street.
- Preserve a continuous central corridor from entrance to endpoint.
- Insert small dining pockets between selected vendors.
- Suspend lighting across the street at irregular intervals.
- Use a gate or prominent sign as the terminal focus.

validation_notes:
- Check hanging-prop pivots before applying height offsets.
- Ensure stalls, furniture, and decorations do not block circulation.
\end{lstlisting}
\end{samepage}

\section{Additional Qualitative Results}
\begin{figure*}[t]
    \centering
    \includegraphics[width=0.98\textwidth]{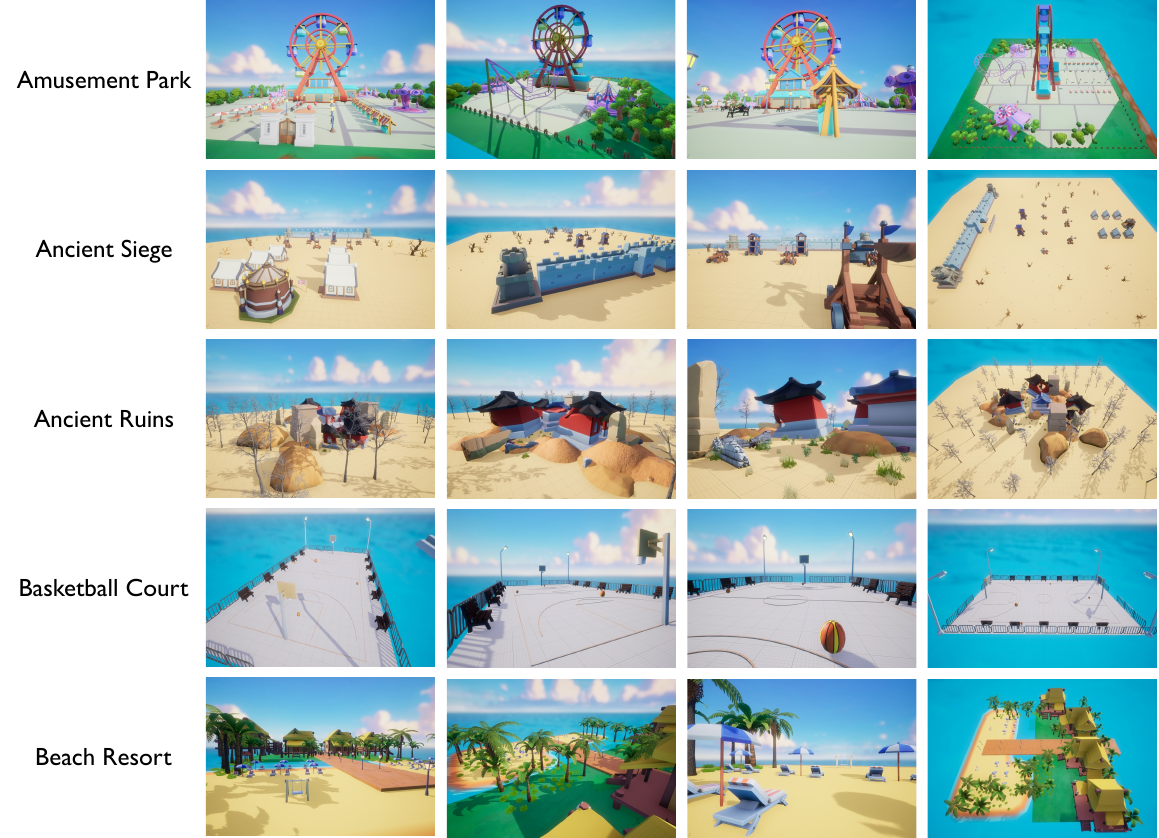}
    \caption{Additional qualitative results in Engine~A (set 1). From top to
    bottom: Amusement Park, Ancient Siege, Ancient Ruins, Basketball Court,
    and Beach Resort. Each row shows four views of the same generated scene.}
    \label{fig:supp-engine-a-results-1}
\end{figure*}

\begin{figure*}[t]
    \centering
    \includegraphics[width=0.98\textwidth]{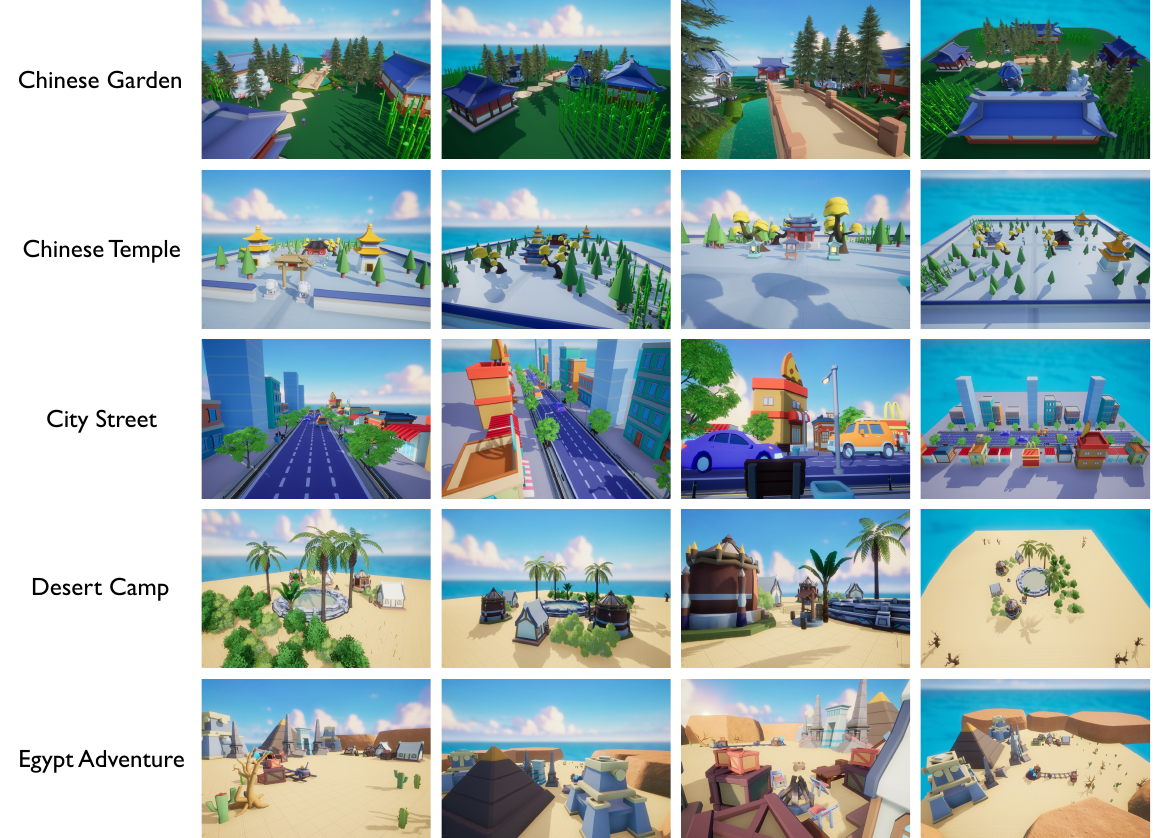}
    \caption{Additional qualitative results in Engine~A (set 2). From top to
    bottom: Chinese Garden, Chinese Temple, City Street, Desert Camp, and
    Egypt Adventure. Each row shows four views of the same generated scene.}
    \label{fig:supp-engine-a-results-2}
\end{figure*}

\begin{figure*}[t]
    \centering
    \includegraphics[width=0.98\textwidth]{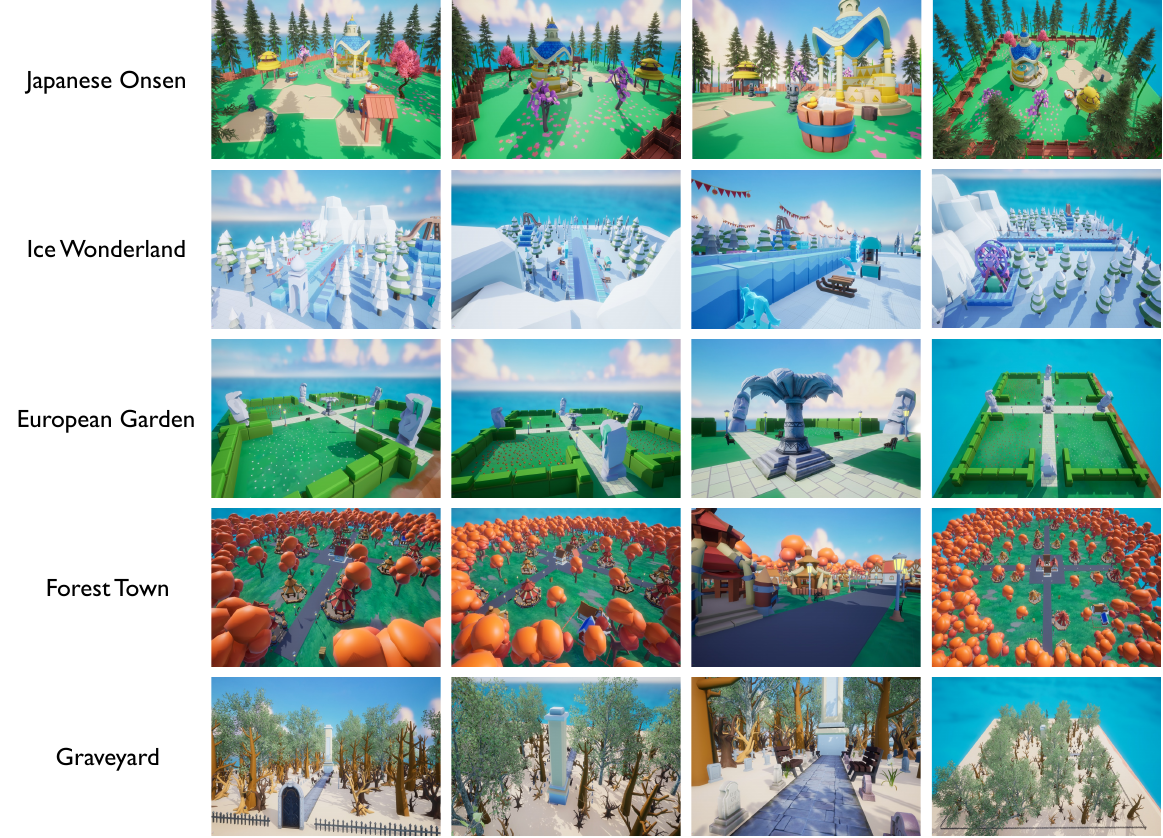}
    \caption{Additional qualitative results in Engine~A (set 3). From top to
    bottom: Japanese Onsen, Ice Wonderland, European Garden, Forest Town, and
    Graveyard. Each row shows four views of the same generated scene.}
    \label{fig:supp-engine-a-results-3}
\end{figure*}

\begin{figure*}[t]
    \centering
    \includegraphics[width=0.98\textwidth]{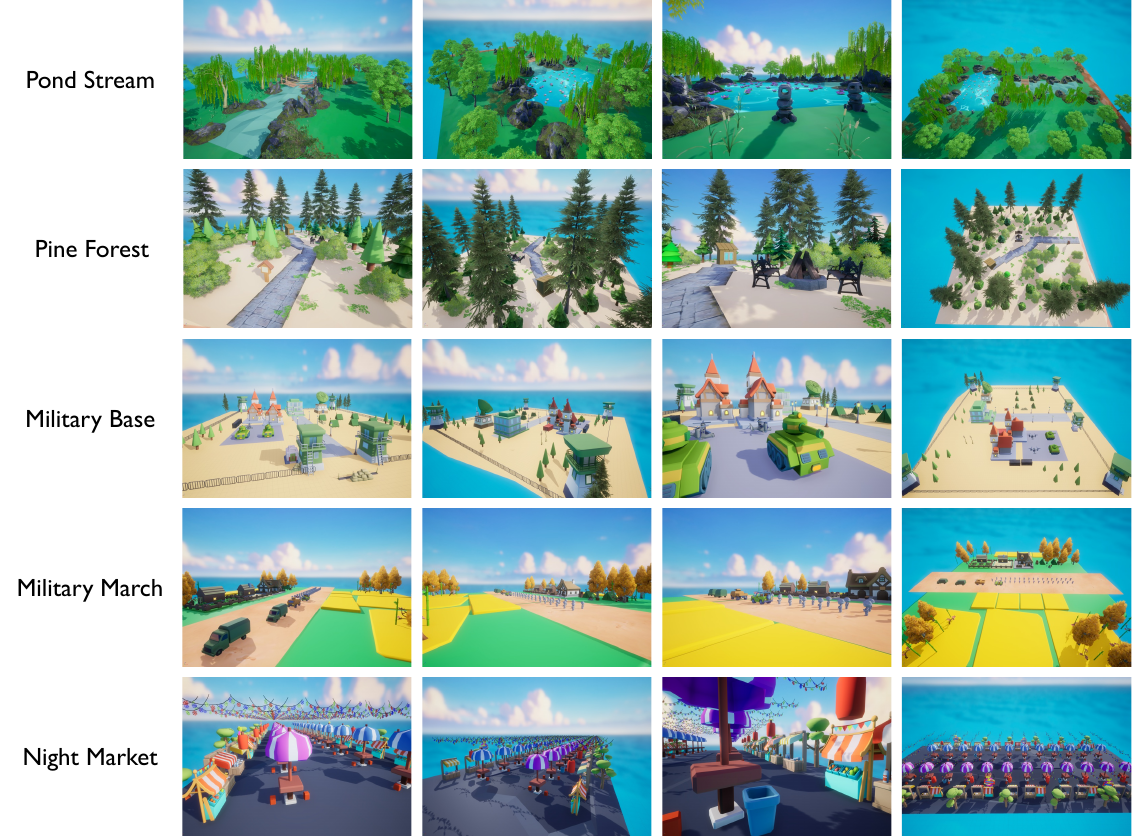}
    \caption{Additional qualitative results in Engine~A (set 4). From top to
    bottom: Pond Stream, Pine Forest, Military Base, Military March, and Night
    Market. Each row shows four views of the same generated scene.}
    \label{fig:supp-engine-a-results-4}
\end{figure*}

\begin{figure*}[t]
    \centering
    \includegraphics[width=0.98\textwidth]{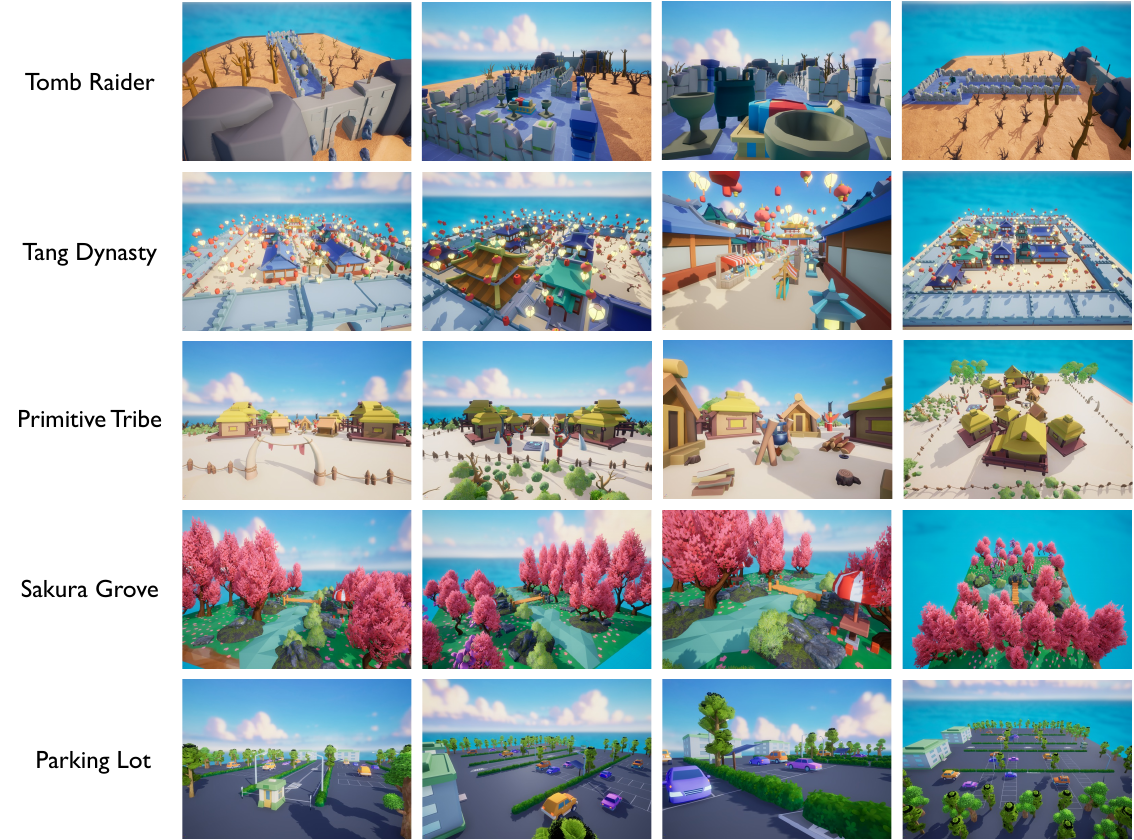}
    \caption{Additional qualitative results in Engine~A (set 5). From top to
    bottom: Tomb Raider, Tang Dynasty, Primitive Tribe, Sakura Grove, and
    Parking Lot. Each row shows four views of the same generated scene.}
    \label{fig:supp-engine-a-results-5}
\end{figure*}

\begin{figure*}[t]
    \centering
    \includegraphics[width=0.98\textwidth]{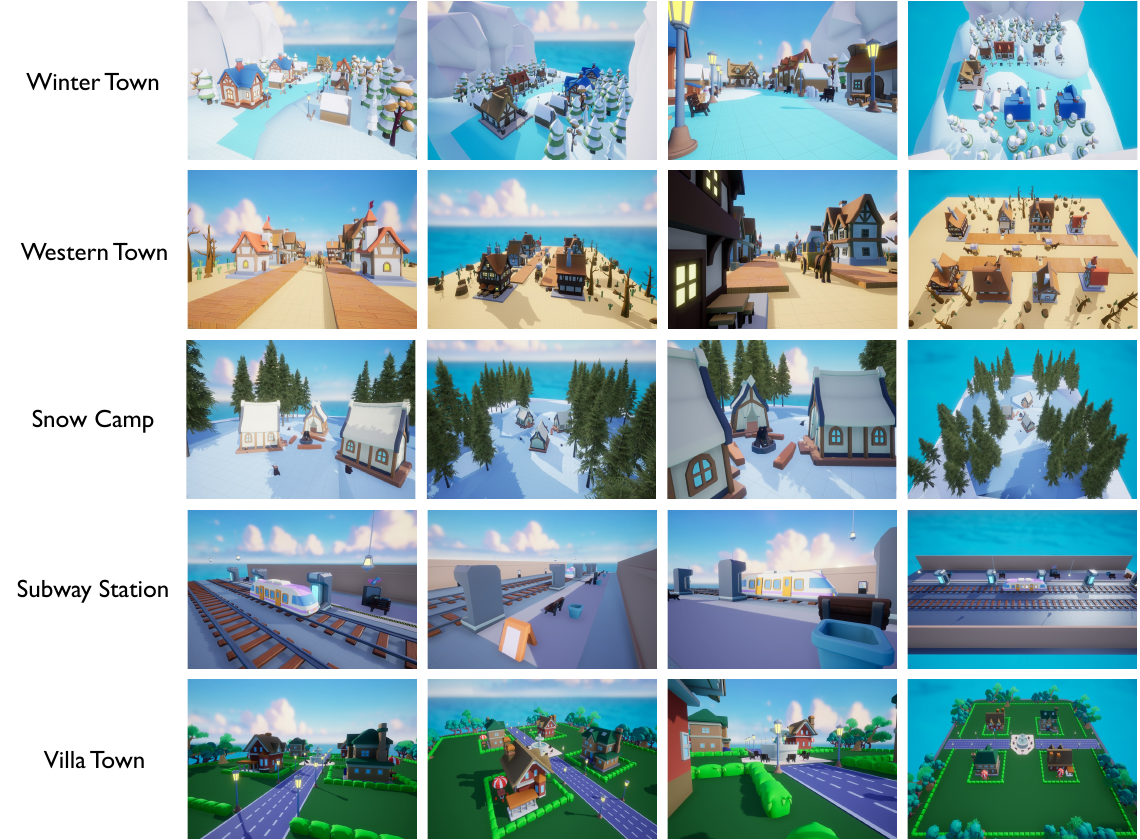}
    \caption{Additional qualitative results in Engine~A (set 6). From top to
    bottom: Winter Town, Western Town, Snow Camp, Subway Station, and Villa
    Town. Each row shows four views of the same generated scene.}
    \label{fig:supp-engine-a-results-6}
\end{figure*}

\begin{figure*}[t]
    \centering
    \includegraphics[width=0.98\textwidth]{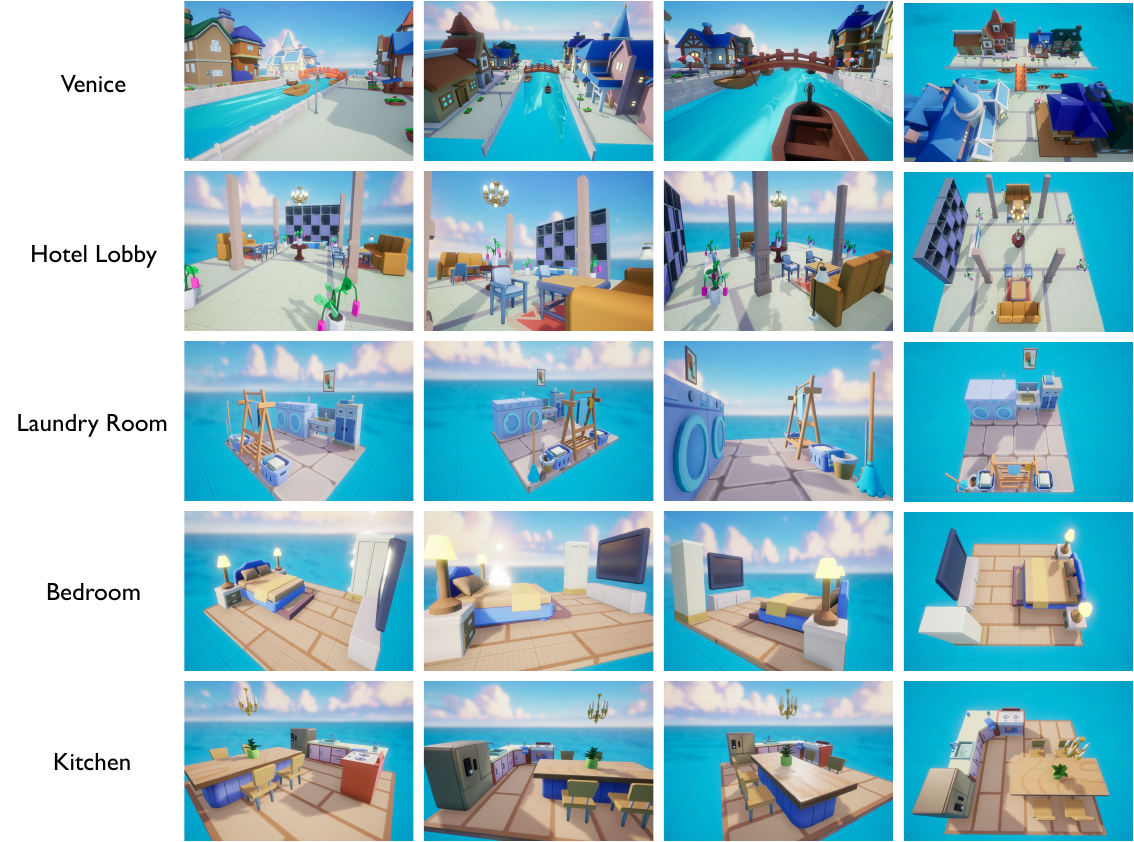}
    \caption{Additional qualitative results in Engine~A (set 7). From top to
    bottom: Venice, Hotel Lobby, Laundry Room, Bedroom, and Kitchen. Each row
    shows four views of the same generated scene.}
    \label{fig:supp-engine-a-results-7}
\end{figure*}

\begin{figure*}[t]
    \centering
    \includegraphics[width=0.98\textwidth]{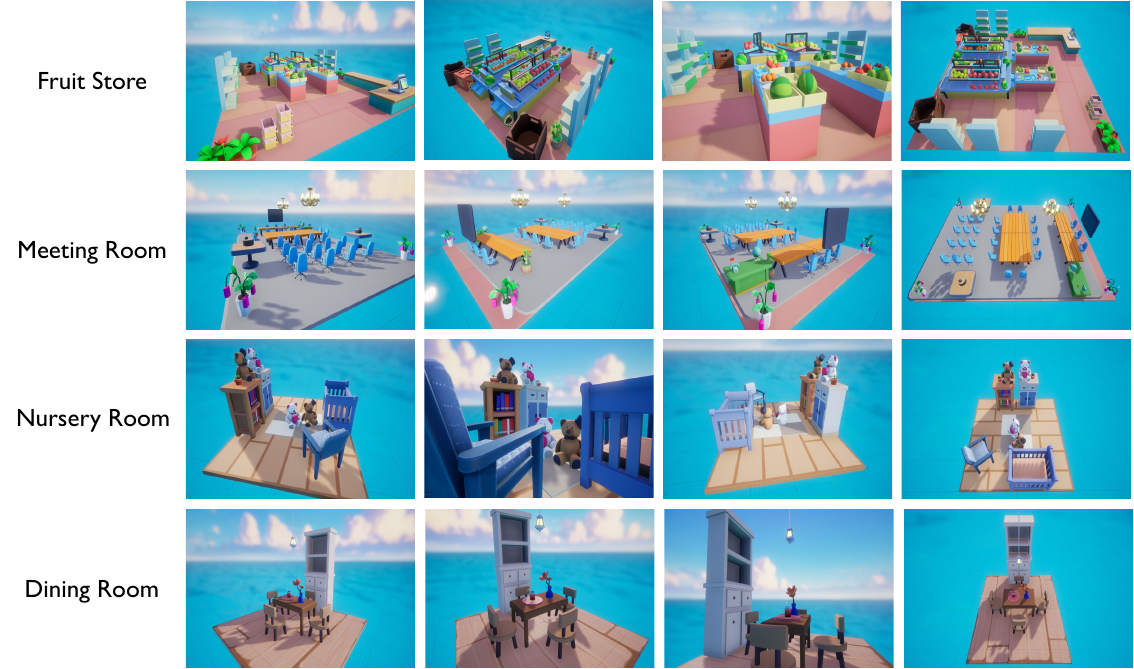}
    \caption{Additional qualitative results in Engine~A (set 8). From top to
    bottom: Fruit Store, Meeting Room, Nursery Room, and Dining Room. Each row
    shows four views of the same generated scene.}
    \label{fig:supp-engine-a-results-8}
\end{figure*}

\begin{figure*}[t]
    \centering
    \includegraphics[width=0.98\textwidth]{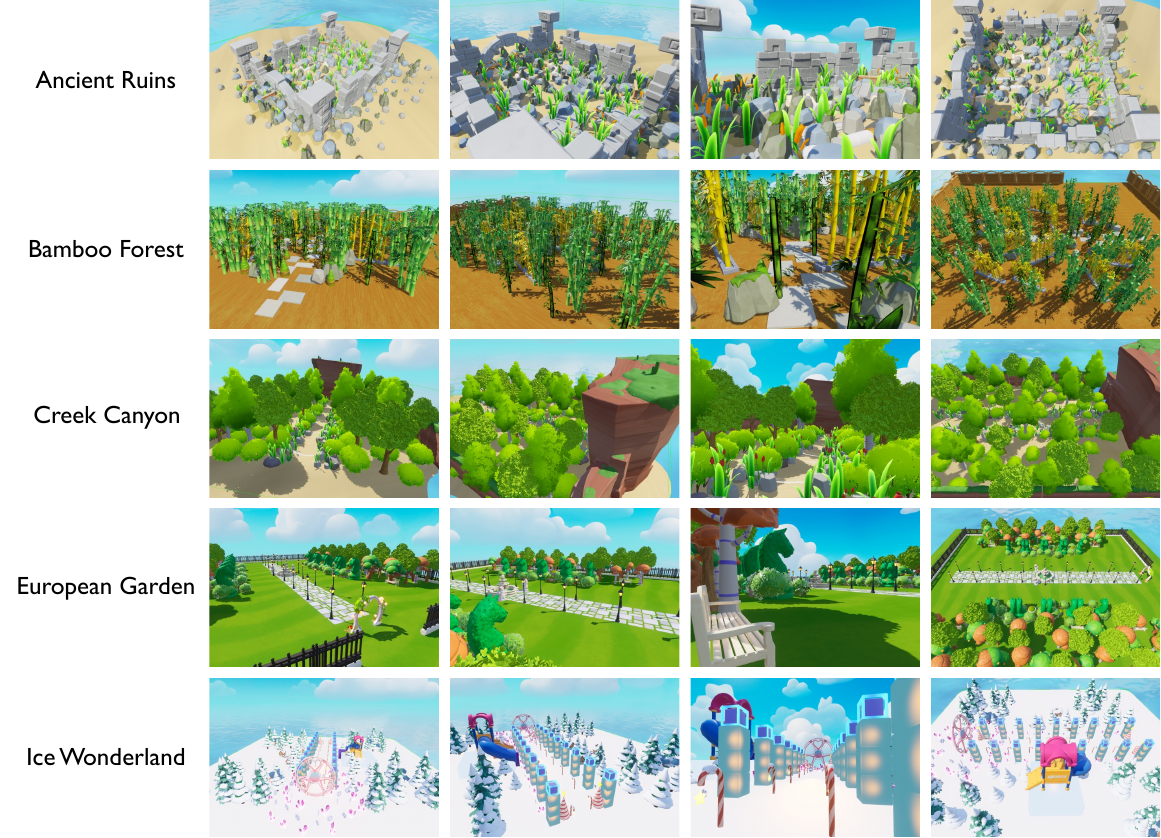}
    \caption{Additional qualitative results in Engine~B (set 1). From top to
    bottom: Ancient Ruins, Bamboo Forest, Creek Canyon, European Garden, and
    Ice Wonderland. Each row shows four views of the same generated scene.}
    \label{fig:supp-engine-b-results-1}
\end{figure*}

\begin{figure*}[t]
    \centering
    \includegraphics[width=0.98\textwidth]{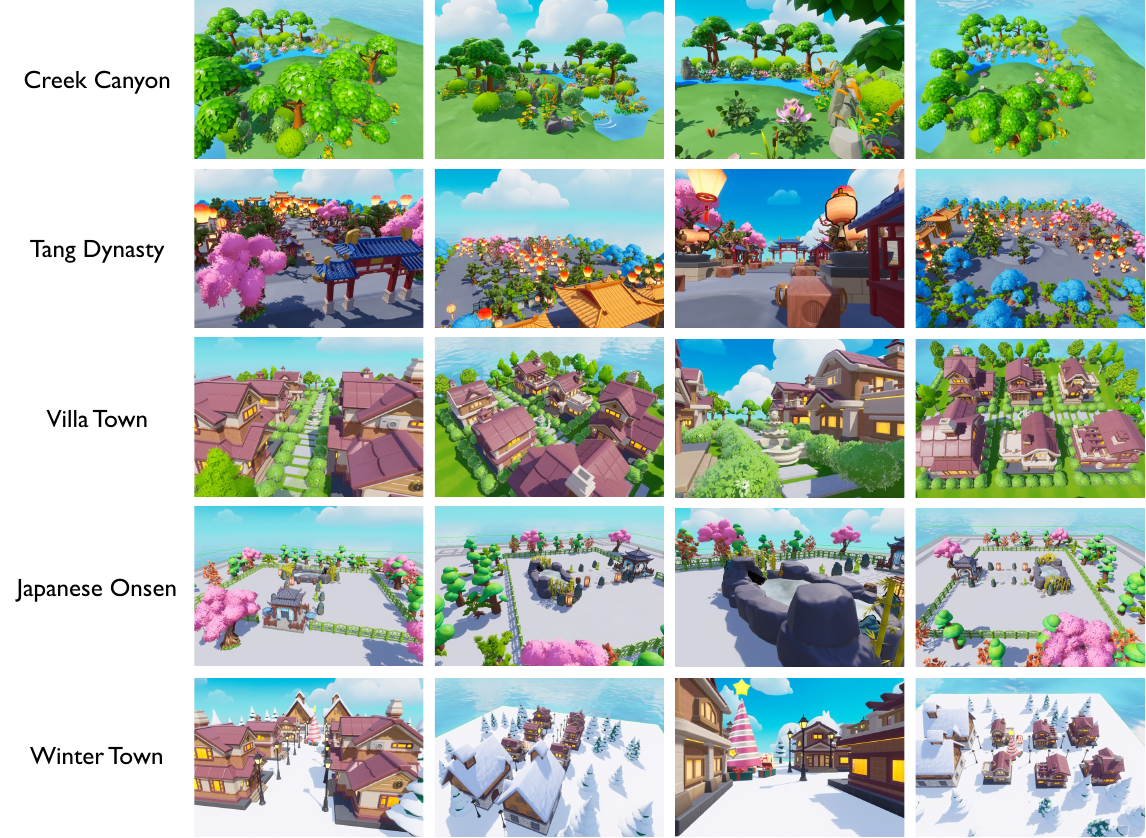}
    \caption{Additional qualitative results in Engine~B (set 2). From top to
    bottom: Creek Canyon, Tang Dynasty, Villa Town, Japanese Onsen, and Winter Town. Each
    row shows four views of the same generated scene.}
    \label{fig:supp-engine-b-results-2}
\end{figure*}

Figures~\ref{fig:supp-engine-a-results-1}--\ref{fig:supp-engine-a-results-8}
show additional generations in Engine~A, and
Figures~\ref{fig:supp-engine-b-results-1}--\ref{fig:supp-engine-b-results-2}
show additional generations in Engine~B.

\ifdefined\SceneAgentCombined
\else
\clearpage
\bibliographystyle{ieeenat_fullname}
\bibliography{main}
\fi

\end{document}